\PassOptionsToPackage{hyphens}{url}

\documentclass{ieeeaccess}
\def\BibTeX{{\rm B\kern-.05em{\sc i\kern-.025em b}\kern-.08em
		T\kern-.1667em\lower.7ex\hbox{E}\kern-.125emX}}
\ifCLASSINFOpdf

\else\usepackage{biblatex}
\fi

\usepackage{url}
\usepackage{soul}

\usepackage{epstopdf, epsfig} 
\usepackage{amssymb}
\usepackage{cite}
\usepackage{graphicx}
\usepackage{makecell}
\usepackage{tabularx,booktabs}
\usepackage{subfig}
\usepackage{eurosym}
\usepackage{xfrac}
\usepackage{hyperref}
\usepackage{cleveref}

\usepackage{xcolor} 
\usepackage{color,soul}

\usepackage{ctable}
\usepackage{multirow}
\usepackage{cleveref}
\usepackage{xcolor} 
\usepackage{color,soul}

\newcommand\T{\rule{0pt}{2.6ex}}        
\newcommand\B{\rule[-1.2ex]{0pt}{0pt}}  
\DeclareRobustCommand{\hsout}[1]{\texorpdfstring{\sout{#1}}{#1}}
\newif\ifversionwithcorrections      \versionwithcorrectionstrue %
\ifversionwithcorrections
\newcommand{\removedtext}[1]{{\colorbox{yellow!30}{\st{#1}}}}
\newcommand{\rremovedtext}[1]{{\colorbox{yellow!30}{\hsout{#1}}}}

\else
\newcommand{\removedtext}[1]{}
\newcommand{\rremovedtext}[1]{}

\fi

\newcolumntype{Y}{>{\centering\arraybackslash}X}

\begin{document}

\history{Accepted March 22, 2020}
\doi{10.1109/ACCESS.2020.2983149}

\title{A Survey of Autonomous Driving: \textit{Common Practices and Emerging Technologies}}

\author{\uppercase{Ekim~Yurtsever}\authorrefmark{1}, \IEEEmembership{Member, IEEE},
	\uppercase{Jacob~Lambert \authorrefmark{1}, Alexander~Carballo} \authorrefmark{1},
	\IEEEmembership{Member, IEEE}, AND \uppercase{Kazuya Takeda} \authorrefmark{1, 2}, \IEEEmembership{Senior Member, IEEE}}
\address[1]{Nagoya University, Furo-cho, Nagoya, 464-8603, Japan}
\address[2]{Tier4 Inc. Nagoya, Japan}

\corresp{Corresponding author: Ekim Yurtsever (e-mail: ekimyurtsever@gmail.com).}

%
%


	
%
%

\begin{abstract}

Automated driving systems (ADSs) promise a safe, comfortable and efficient driving experience. However, fatalities involving vehicles equipped with ADSs are on the rise. The full potential of ADSs cannot be realized unless the robustness of state-of-the-art is improved further. This paper discusses unsolved problems and surveys the technical aspect of automated driving. Studies regarding present challenges, high-level system architectures, emerging methodologies and core functions including localization, mapping, perception, planning, and human machine interfaces, were thoroughly reviewed. Furthermore, many state-of-the-art algorithms were implemented and compared on our own platform in a real-world driving setting. The paper concludes with an overview of available datasets and tools for ADS development.


 
	
\end{abstract}


\IEEEpeerreviewmaketitle

%

\titlepgskip=-15pt

\maketitle

\section{Introduction}\label{sec_intro}




\IEEEPARstart{A}{ccording} to a recent technical report by the National Highway Traffic Safety Administration (NHTSA), 94\% of road accidents are caused by human errors \cite{singh2015critical}. Against this backdrop, Automated Driving Systems (ADSs) are being developed with the promise of preventing accidents, reducing emissions, transporting the mobility-impaired and reducing driving related stress \cite{crayton2017autonomous}. If widespread deployment can be realized, annual social benefits of ADSs are projected to reach nearly \$800 billion by 2050 through congestion mitigation, road casualty reduction, decreased energy consumption and increased productivity caused by the reallocation of driving time \cite{montgomery2018america}.  

The accumulated knowledge in vehicle dynamics, breakthroughs in computer vision caused by the advent of deep learning \cite{krizhevsky2012imagenet} and availability of new sensor modalities, such as lidar \cite{schwarz2010lidar}, catalyzed ADS research and industrial implementation. Furthermore, an increase in public interest and market potential precipitated the emergence of ADSs with varying degrees of automation. However, robust automated driving in urban environments has not been achieved yet \cite{hecker2018end}. Accidents caused by immature systems \cite{lavrinc2014bad, davies2016google, mcfarland2016, lee2019} undermine trust, and furthermore, cost lives. As such, a thorough investigation of unsolved challenges and the state-of-the-art is deemed necessary here.   

Eureka Project PROMETHEUS \cite{eureka45} was carried out in Europe between 1987-1995, and it was one of the earliest major automated driving studies. The project led to the development of VITA II by Daimler-Benz, which succeeded in automatically driving on highways \cite{ulmer1994vita}. DARPA Grand Challenge, organized by the US Department of Defense in 2004, was the first major automated driving competition where all of the attendees failed to finish the 150-mile off-road parkour. The difficulty of the challenge was in the rule that no human intervention at any level was allowed during the finals. Another similar DARPA Grand Challenge was held in 2005. This time five teams managed to complete the off-road track without any human interference \cite{buehler20072005}. 

\begin{table*}
	\caption{Comparison of ADS related survey papers}
	\label{table_survey_comparision}
	\begin{tabularx}{\textwidth}{@{}cYcYcYccYYc@{}}
		\specialrule{.1em}{.05em}{.05em} 
		\textbf{Related work} &  \multicolumn{10}{c}{\textbf{Survey coverage}}\\
		\hline
		& Connected systems & End-to-end & Localization  & Perception & Assessment & Planning & Control & HMI & Datasets \& software & Implementation  \\ 
		
		\cite{feng2019deep}  & - &- & - & \checkmark & - & - & - & - &-  & - \\
		\cite{levinson2011towards}   & - &- & \checkmark & \checkmark & - & \checkmark & \checkmark & - &-  & \checkmark \\
		\cite{campbell2010autonomous}  &  - & - & \checkmark & \checkmark & - & \checkmark& -& - & - & - \\
		\cite{kuutti2018survey}  &  - & - & \checkmark & - & - & - & -& - & - & - \\
		\cite{paden2016survey}  & - & - & - & - & - & \checkmark& \checkmark & - & - & - \\
		\cite{gonzalez2016review}  & - & - & - & - & - & \checkmark& - & - & - & - \\
		\cite{van2018autonomous}   & \checkmark & \checkmark & \checkmark & \checkmark & - & -& - & - & - & - \\
		\cite{bresson2017simultaneous}  & \checkmark & - & \checkmark & - & - & - & - & - & \checkmark & - \\
		\cite{abboud2016interworking}  & \checkmark & - & - & - & - &-& - & - & - & - \\ 
		\cite{badue2019self}  & - & - & \checkmark & \checkmark & - & \checkmark &\checkmark& - & - & \checkmark\\
		\cite{schwarting2018planning} & - &  \checkmark& - & - & \checkmark  & \checkmark & \checkmark & - & - & - \\
		\cite{lefevre2014survey} & - &  - & - & - & \checkmark  & \checkmark & - & - & - & - \\
		\textbf{Ours} &  \checkmark & \checkmark & \checkmark & \checkmark &  \checkmark & \checkmark & - & \checkmark & \checkmark & \checkmark \\
		\specialrule{.1em}{.05em}{.05em}
		
	\end{tabularx}
	\vspace{-0.5cm}
\end{table*}

Fully automated driving in urban scenes was seen as the biggest challenge of the field since the earliest attempts. During DARPA Urban Challenge \cite{buehler2009darpa}, held in 2007, many different research groups around the globe tried their ADSs in a test environment that was modeled after a typical urban scene. Six teams managed to complete the event. Even though this competition was the biggest and most significant event up to that time, the test environment lacked certain aspects of a real-world urban driving scene such as pedestrians and cyclists. Nevertheless, the fact that six teams managed to complete the challenge attracted significant attention. After DARPA Urban Challenge, several more automated driving competitions such as \cite{broggi2012vislab,broggi2015proud,cerri2011computer, englund2016grand} were held in different countries.   

Common practices in system architecture have been established over the years. Most of the ADSs divide the massive task of automated driving into subcategories and employ an array of sensors and algorithms on various modules. More recently, end-to-end driving started to emerge as an alternative to modular approaches.  Deep learning models have become dominant in many of these tasks \cite{mcallister2017concrete}. 

The Society of Automotive Engineers (SAE) refers to hardware-software systems that can execute dynamic driving tasks (DDT) on a sustainable basis as ADS \cite{sae2016}. There are also vernacular alternative terms such as "autonomous driving" and "self-driving car" in use. Nonetheless, despite being commonly used, SAE advices not to use them as these terms are unclear and misleading. In this paper we follow SAE's convention.

The present paper attempts to provide a structured and comprehensive overview of state-of-the-art automated driving related hardware-software practices. Moreover, emerging trends such as end-to-end driving and connected systems are discussed in detail. There are overview papers on the subject, which covered several core functions \cite{levinson2011towards,campbell2010autonomous}, and which concentrated only on the motion planning aspect \cite{paden2016survey, gonzalez2016review}. However, a survey that covers: present challenges, available and emerging high-level system architectures, individual core functions such as localization, mapping, perception, planning, vehicle control, and human-machine interface altogether does not exist. The aim of this paper is to fill this gap in the literature with a thorough survey. In addition, a detailed summary of available datasets, software stacks, and simulation tools is presented here. Another contribution of this paper is the detailed comparison and analysis of alternative approaches through implementation. We implemented some state-of-the-art algorithms in our platform using open-source software. Comparison of existing overview papers and our work is shown in Table \ref{table_survey_comparision}. 



The remainder of this paper is written in eight sections. Section \ref{sec_challenges} is an overview of present challenges. Details of automated driving system components and architectures are given in Section \ref{sec_components}. Section \ref{sec_localization} presents a summary of state-of-the-art localization techniques followed by Section \ref{sec:perception}, an in-depth review of perception models. Assessment of the driving situation and planning are discussed in Section VI and VII respectively. In Section VIII, current trends and shortcomings of human machine interface are introduced. Datasets and available tools for developing automated driving systems are given in Section IX. 

%

\section{Prospects and Challenges}\label{sec_challenges}

\subsection{Social impact}\label{s:social_issues}
Widespread usage of ADSs is not imminent. Yet it is still possible to foresee its potential impact and benefits to a certain degree:  
\begin{enumerate}
	\item \textit{Problems that can be solved}: preventing traffic accidents, mitigating traffic congestions, reducing emissions
	\item \textit{Arising opportunities}: reallocation of driving time, transporting the mobility impaired
	\item \textit{New trends}: consuming Mobility as a Service (MaaS), logistics revolution
\end{enumerate}

Widespread deployment of ADSs can reduce the societal loss caused by erroneous human behavior such as distraction, driving under influence and speeding \cite{montgomery2018america}. 

Globally, the elder group (over 60 years old) is growing faster than the younger groups\cite{desa2017}. Increasing the mobility of elderly with ADSs can have a huge impact on the quality of life and productivity of a large portion of the population. 

A shift from personal vehicle-ownership towards consuming Mobility as a Service (MaaS) is an emerging trend. Currently, ride-sharing has lower costs compared to vehicle-ownership under 1000 km annual mileage \cite{deloitte2019}. The ratio of owned to shared vehicles is expected to be 50:50 by 2030 \cite{fiareion1_2019}. Large scale deployment of ADSs can accelerate this trend.

\subsection{Challenges}\label{s:automation_levels}

ADSs are complicated robotic systems that operate in indeterministic environments. As such, there are myriad scenarios with unsolved issues. This section discusses the high level challenges of driving automation in general. More minute, task-specific details are discussed in corresponding sections.

The Society of Automotive Engineers (SAE) defined five levels of driving automation in \cite{sae2016}. In this taxonomy, \emph{level zero} stands for no automation at all. Primitive driver assistance systems such as adaptive cruise control, anti-lock braking systems and stability control start with \emph{level one} \cite{rajamani2011vehicle}. \emph{Level two} is partial automation to which advanced assistance systems such as emergency braking or collision avoidance \cite{hafner2013cooperative, colombo2012efficient} are integrated. With the accumulated knowledge in the vehicle control field and the experience of the industry, level two automation became a feasible technology. The real challenge starts above this level.

\emph{Level three} is conditional automation; the driver could focus on tasks other than driving during normal operation, however, s/he has to quickly respond to an emergency alert from the vehicle and be ready to take over. In addition, level three ADS operate only in limited operational design domains (ODDs) such as highways. Audi claims to be the first production car to achieve level 3 automation in limited highway conditions \cite{ross2017audi}. However, taking over the control manually from the automated mode by the driver raises another issue. Recent studies \cite{gold2016taking}, \cite{merat2014transition} investigated this problem and found that the takeover situation increases the collision risk with surrounding vehicles. The increased likelihood of an accident during a takeover is a problem that is yet to be solved.

Human attention is not needed in any degree at level four and five. However, \emph{level four} can only operate in limited ODDs where special infrastructure or detailed maps exist. In the case of departure from these areas, the vehicle must stop the trip by automatically parking itself. The fully automated system, \emph{level five}, can operate in any road network and any weather condition. No production vehicle is capable of level four or level five driving automation yet. Moreover, Toyota Research Institute stated that no one in the industry is even close to attaining level five automation \cite{ackerman2017toyota}.     

Level four and above driving automation in urban road networks is an open and challenging problem. The environmental variables, from weather conditions to surrounding human behavior, are highly indeterministic and difficult to predict. Furthermore, system failures lead to accidents: in the Hyundai competition one of the ADSs crashed because of rain \cite{lavrinc2014bad}, Google's ADS hit a bus while lane changing because it failed to estimate the speed of a bus \cite{davies2016google}, and Tesla's Autopilot failed to recognize a white truck and collided with it, killing the driver \cite{mcfarland2016}.

Fatalities \cite{ mcfarland2016, lee2019} caused by immature technology undermine public acceptance of ADSs. According to a recent survey \cite{deloitte2019}, the majority of consumers question the safety of the technology, and want a significant amount of control over the development and use of ADS. On the other hand, extremely cautious ADSs are also making a negative impression \cite{donfro2018}. 

Ethical dilemmas pose another set of challenges. In an inevitable accident situation, how should the system behave \cite{bonnefon2016social}? Experimental ethics were proposed regarding this issue \cite{bonnefon2015autonomous}.

Risk and reliability certification is another task yet to be solved. Like in aircraft, ADSs need to be designed with high redundancies that will minimize the chance of a catastrophic failure. Even though there is promising projects in this regard such as DeepTest \cite{tian2018deeptest}, the design-simulation-test-redesign-certification procedure is still not established by the industry nor the rule-makers.



Finally, various optimization goals such as time to reach the destination, fuel efficiency, comfort, and ride-sharing optimization increases the complexity of an already difficult to solve problem. As such, carrying all of the dynamic driving tasks safely under strict conditions outside a well defined, geofenced area remains as an open problem.

\begin{figure}
	\centering\includegraphics[width=1\columnwidth]{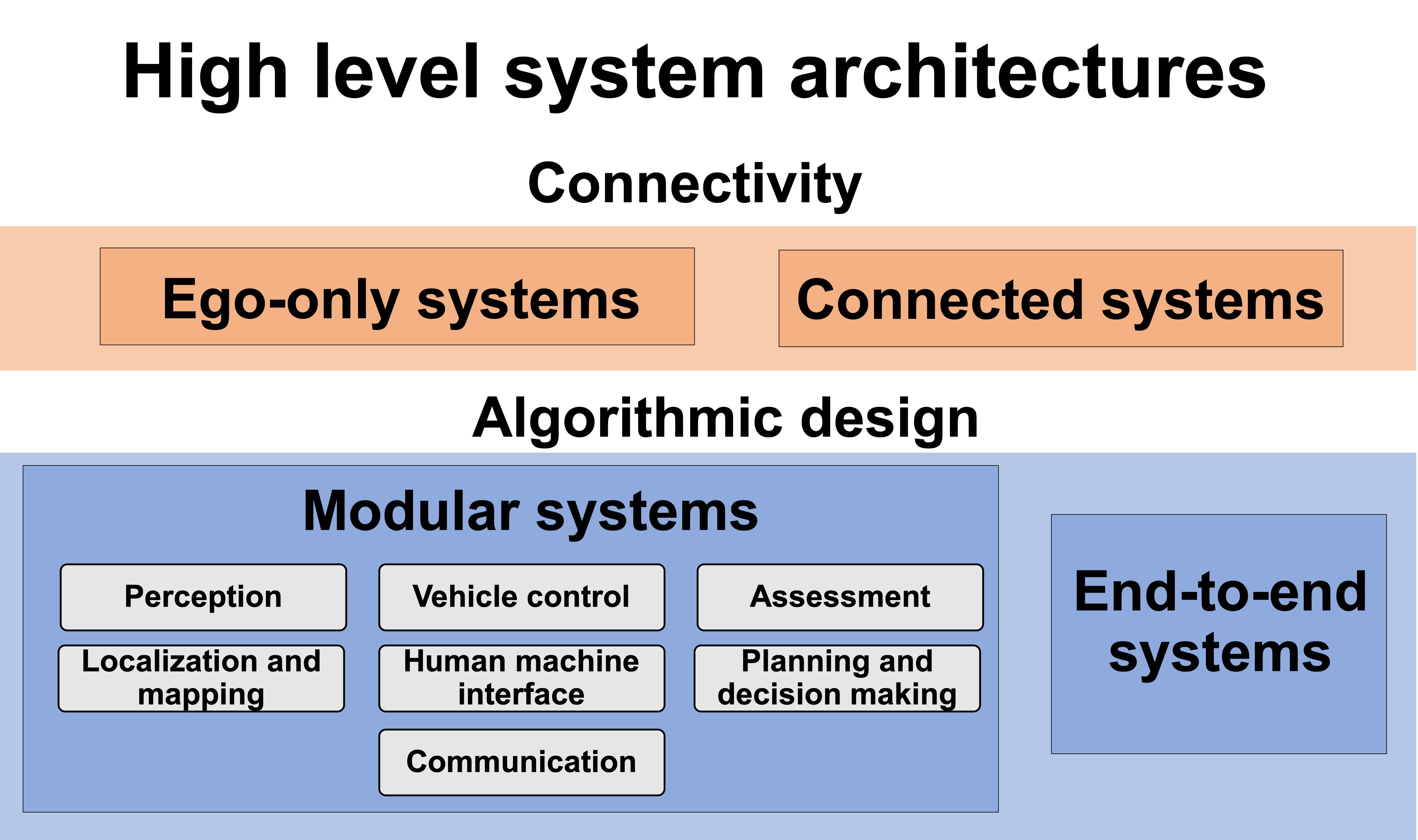}
	\caption{A high level classification of automated driving system architectures}
	\label{fig_high_level}
\end{figure}

\begin{figure*}
	\centering\includegraphics[width=1.5\columnwidth]{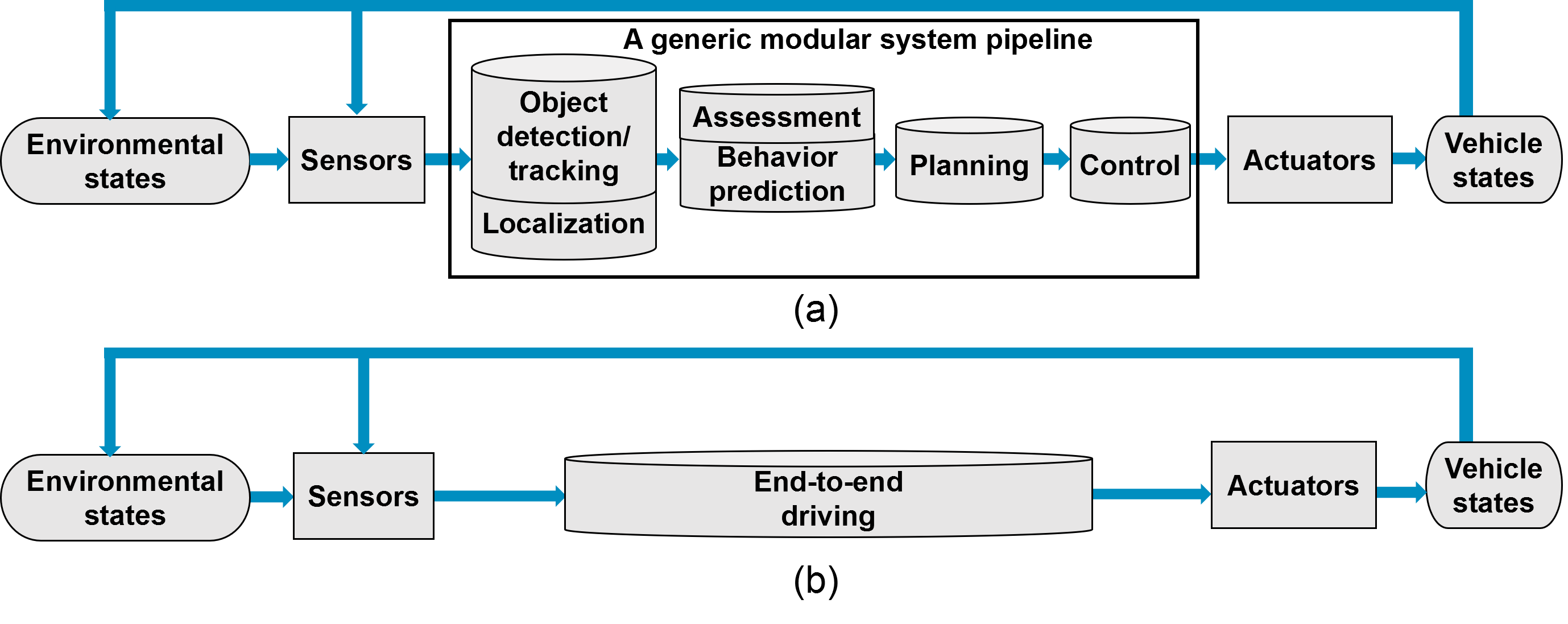}
	\caption{ Information flow diagrams of: (a) a generic modular system, and (b) an end-to-end driving system. }
	\label{fig_end2end}
\end{figure*}

\section{System components and architecture}\label{sec_components}



\subsection{System architecture}
Classification of system architectures is shown in Figure \ref{fig_high_level}.
ADSs are designed either as standalone, ego-only systems \cite{urmson2008autonomous, levinson2011towards} or connected multi-agent systems \cite{gerla2014internet,lee2016internet,amadeo2016information}. Furthermore, these design philosophies are realized with two alternative approaches: modular \cite{urmson2008autonomous, levinson2011towards, wei2013towards, broggi2013extensive, maddern20171, akai2017autonomous, guizzo2011google, somerville2018uber, ziegler2014making, apolloauto} or end-to-end driving \cite{chen2015deepdriving, pomerleau1989alvinn, muller2006off, bojarski2016end, xu2017end, sallab2017deep, kendall2018learning, baluja1996evolution, koutnik2013evolving}.   
 
 



\subsubsection{Ego-only systems}

The ego-only approach is to carry all of the necessary automated driving operations on a single self-sufficient vehicle at all times, whereas a connected ADS may or may not depend on other vehicles and infrastructure elements given the situation. Ego-only is the most common approach amongst the state-of-the-art ADSs  \cite{urmson2008autonomous, levinson2011towards, wei2013towards, broggi2013extensive, maddern20171, akai2017autonomous, guizzo2011google, somerville2018uber, somerville2018uber, ziegler2014making, apolloauto}. We believe this is due to the practicality of having a self-sufficient platform for development and the additional challenges of connected systems. 


\subsubsection{Modular systems}

Modular systems, referred as the mediated approach in some works \cite{chen2015deepdriving}, are structured as a pipeline of separate components linking sensory inputs to actuator outputs \cite{mcallister2017concrete}. Core functions of a modular ADS can be summarized as: localization and mapping, perception, assessment, planning and decision making, vehicle control, and human-machine interface. Typical pipelines \cite{urmson2008autonomous, levinson2011towards, wei2013towards, broggi2013extensive, maddern20171, akai2017autonomous, guizzo2011google, somerville2018uber, somerville2018uber, ziegler2014making, apolloauto} start with feeding raw sensor inputs to localization and object detection modules, followed by scene prediction and decision making. Finally, motor commands are generated at the end of the stream by the control module \cite{mcallister2017concrete, behere2015functional}. 


Developing individual modules separately divides the challenging task of automated driving into an easier-to-solve set of problems \cite{chi2017deep}. These sub-tasks have their corresponding literature in robotics \cite{laumond1998robot}, computer vision \cite{jain1995machine} and vehicle dynamics \cite{rajamani2011vehicle}, which makes the accumulated know-how and expertise directly transferable. This is a major advantage of modular systems. In addition, functions and algorithms can be integrated or built upon each other in a modular design. E.g, a safety constraint \cite{anderson2012constraint} can be implemented on top of a sophisticated planning module to force some hard-coded emergency rules without modifying the inner workings of the planner. This enables designing redundant but reliable architectures.





The major disadvantages of modular systems are being prone to error propagation \cite{mcallister2017concrete} and over-complexity. In the unfortunate Tesla accident, an error in the perception module in the form of a misclassification of a white trailer as sky, propagated down the pipeline until failure, causing the first ADS related fatality \cite{tian2018deeptest}.  




\subsubsection{End-to-end driving}

End-to-end driving, referred as direct perception in some studies \cite{chen2015deepdriving}, generate ego-motion directly from sensory inputs. Ego-motion can be either the continuous operation of steering wheel and pedals or a discrete set of actions, e.g, acceleration and turning left. There are three  main approaches for end-to-end driving: direct supervised deep learning \cite{pomerleau1989alvinn, muller2006off, bojarski2016end, xu2017end, chen2015deepdriving}, neuroevolution \cite{koutnik2013evolving, baluja1996evolution} and the more recent deep reinforcement learning \cite{sallab2017deep, kendall2018learning}. The flow diagram of a generic end-to-end driving system is shown in Figure \ref{fig_end2end} and comparison of the approaches is given in Table \ref{table_end2end}.   


\begin{table}
	\footnotesize
	\renewcommand{\arraystretch}{1.3}
	\caption{Common end-to-end driving approaches}
	\label{table_end2end}
	\centering
	\begin{tabularx}{\columnwidth}{@{}cYY@{}}
		\specialrule{.1em}{.05em}{.05em} 
		\B \T	
		\thead{\textbf{Related} \\ \textbf{works}}  & \textbf{Learning/training strategy} & \textbf{Pros/cons} \T \\
		\hline
		\cite{pomerleau1989alvinn, muller2006off, bojarski2016end, xu2017end, chen2015deepdriving} & Direct supervised deep learning & Imitates the target data: usually a human driver. Can be trained offline. Poor generalization performance.   \\
		\cite{sallab2017deep, kendall2018learning} & Deep reinforcement learning & Learns the optimum way of driving. Requires online interaction. Urban driving has not been achieved yet \\
		\cite{koutnik2013evolving, baluja1996evolution} & Neuroevolution & No backpropagation. Requires online interaction. Real world driving has not been achieved yet.  \\ 	
		\specialrule{.1em}{.05em}{.05em} 
	\end{tabularx}
	\vspace{-0.5cm}
\end{table}

The earliest end-to-end driving attempt dates back to ALVINN \cite{pomerleau1989alvinn}, where a 3-layer fully connected network was trained to output the direction that the vehicle should follow. An end-to-end driving system for off-road driving was introduced in \cite{muller2006off}. With the advances in artificial neural network research, deep convolutional and temporal networks became feasible for automated driving tasks.  A deep convolutional neural network that takes image as input and outputs steering was proposed in \cite{bojarski2016end}. A spatiotemporal network, an FCN-LSTM architecture, was developed for predicting ego-vehicle motion in \cite{xu2017end}. DeepDriving is another convolutional model that tries to learn a set of discrete perception indicators from the image input \cite{chen2015deepdriving}. This approach is not entirely end-to-end though, the proper driving actions in the perception indicators have to be generated by another module. All of the mentioned methods follow direct supervised training strategies. As such, ground truth is required for training. Usually, the ground truth is the ego-action sequence of an expert human driver and the network learns to imitate the driver. This raises an import design question: should the ADS drive like a human? 

A novel deep reinforcement learning model, Deep Q Networks (DQN), combined reinforcement learning with deep learning \cite{mnih2015human}. In summary, the goal of the network is to select a set of actions that maximize cumulative future rewards. A deep convolutional neural network was used to approximate the optimal action reward function. Actions are generated first with random initialization. Then, the network adjust its parameters with experience instead of direct supervised learning. An automated driving framework using DQN was introduced in \cite{sallab2017deep}, where the network was tested in a simulation environment. The first real world run with DQN was achieved in a countryside road without traffic \cite{kendall2018learning}. DQN based systems do not imitate the human driver, instead, they learn the optimum way of driving. 

Neuroevolution refers to using evolutionary algorithms to train artificial neural networks \cite{floreano2008neuroevolution}. End-to-end driving with neuroevolution is not popular as DQN and direct supervised learning. To the best of our knowledge, real world end-to-end driving with neuroevolution is not achieved yet. However, some promising simulation results were obtained \cite{baluja1996evolution, koutnik2013evolving}. ALVINN was trained with neuroevolution and outperformed the direct supervised learning version \cite{baluja1996evolution}. A RNN was trained with neuroevolution in \cite{koutnik2013evolving} using a driving simulator. The biggest advantage of neuroevolution is the removal of backpropagation, hence, the need for direct supervision. 


End-to-end driving is promising, however it has not been implemented in real-world urban scenes yet, except limited demonstrations. The biggest shortcomings of end-to-end driving in general are the lack of hard coded safety measures and interpretability \cite{chi2017deep}. In addition, DQN and neuroevolution has one major disadvantage over direct supervised learning: these networks must interact with the environment online and fail repeatedly to learn the desired behavior. On the contrary, direct supervised networks can be trained offline with human driving data, and once the training is done, the system is not expected to fail during operation.  




\subsubsection{Connected systems}

There is no operational connected ADS in use yet, however, some researchers believe this emerging technology will be the future of driving automation \cite{gerla2014internet,lee2016internet,amadeo2016information}. With the use of Vehicular Ad hoc NETwork (VANETs), the basic operations of automated driving can be distributed amongst agents. V2X is a term that stands for ``vehicle to everything." From mobile devices of pedestrians to stationary sensors on a traffic light, an immense amount of data can be accessed by the vehicle with V2X \cite{abboud2016interworking}. By sharing detailed information of the traffic network amongst peers \cite{cheng2011infotainment}, shortcomings of the ego-only platforms such as sensing range, blind spots, and computational limits may be eliminated. More V2X applications that will increase safety and traffic efficiency are expected to emerge in the foreseeable future \cite{wang2019survey}. 


\begin{table*}[!t]
	\caption{Exteroceptive  sensors}
	\label{table_sensors_comp} 
	{\begin{tabularx}{\textwidth}{@{}cYYYYYYYY@{}}
			\specialrule{.1em}{.05em}{.05em} 
			
			Modality & \thead{Affected by \\ Illumination}& \thead{Affected by \\  weather} & Color & Depth & Range & Accuracy & Size & Cost  \\ 
			\hline
			Lidar & - & \checkmark & - & \checkmark & \thead{medium \\ ($<200$m) } &  high & large* & high*  \\
			Radar & -  & - & - & \checkmark & high & medium & small & medium  \\
			Ultrasonic  & - & -  & -  & \checkmark & short & low  & small & low \\
			Camera & \checkmark & \checkmark & \checkmark & - & - & - & smallest & lowest  \\
			Stereo Camera & \checkmark & \checkmark & \checkmark & \checkmark & \thead{medium \\ ($<100$m) } & low & medium & low  \\
			Flash Camera \cite{jang2017design} & \checkmark & \checkmark & \checkmark & \checkmark & \thead{medium \\ ($<100$m) } & low & medium & low  \\
			Event Camera \cite{maqueda2018event}& limited & \checkmark & - & - & - & - & smallest & low  \\
			Thermal Camera \cite{fries2015autonomous, ha2017mfnet} & - & \checkmark & - & - & - & - & smallest & low  \\
			\specialrule{.1em}{.05em}{.05em}
			\multicolumn{9}{l}{* Cost, size and weight of lidars started to decrease recently \cite{lee-lidar2019}	}
	\end{tabularx}}{}
\end{table*}

VANETs can be realized in two different ways: conventional IP based networking and Information-Centric Networking (ICN) \cite{gerla2014internet}. For vehicular applications, lots of data have to be distributed amongst agents with intermittent and in less than ideal connections while maintaining high mobility \cite{amadeo2016information}. Conventional IP-host based Internet protocol cannot function properly under these conditions. On the other hand, in information-centric networking,  vehicles stream query messages to an area instead of a direct address and they accept corresponding responses from any sender \cite{lee2016internet}. Since vehicles are highly mobile and dispersed on the road network, the identity of the information source becomes less relevant. In addition, local data often carries more crucial information for immediate driving tasks such as avoiding a rapidly approaching vehicle on a blind spot.   

Early works, such as the CarSpeak system \cite{kumar2012carspeak}, proved that vehicles can utilize each other's sensors and use the shared information to execute some dynamic driving tasks. However, without reducing huge amounts of continuous driving data, sharing information between hundreds of thousand vehicles in a city could not become feasible. A semiotic framework that integrates different sources of information and converts raw sensor data into meaningful descriptions was introduced in \cite{yurtsever2018integrating} for this purpose. In \cite{gerla2012vehicular}, the term Vehicular Cloud Computing (VCC) was coined  and the main advantages of it over conventional Internet cloud applications was introduced. Sensors are the primary cause of the difference. In VCC, sensor information is kept on the vehicle and only shared if there is a local query from another vehicle. This potentially saves the cost of uploading/downloading a constant stream of sensor data to the web. Besides, the high relevance of local data increases the feasibility of VCC. Regular cloud computing was compared to vehicular cloud computing and it was reported that VCC is technologically feasible \cite{whaiduzzaman2014survey}. The term ''Internet of Vehicles" (IoV) was proposed for describing a connected ADS \cite{gerla2014internet} and the term ''vehicular fog" was introduced in \cite{lee2016internet}.

Establishing an efficient VANET with thousands of vehicles in a city is a huge challenge.  For an ICN based VANET, some of the challenging topics are security, mobility, routing, naming, caching, reliability and multi-access computing \cite{din2018information}. In summary, even though the potential benefits of a connected system is huge, the additional challenges increase the complexity of the problem to a significant degree. As such, there is no operational connected system yet. 

\subsection{Sensors and hardware}\label{sec:sensors}

State-of-the-art ADSs employ a wide selection of onboard sensors. High sensor redundancy is needed in most of the tasks for robustness and reliability. Hardware units can be categorized into five: exteroceptive sensors for perception, proprioceptive sensors for internal vehicle state monitoring tasks, communication arrays, actuators, and computational units.   


Exteroceptive sensors are mainly used for perceiving the environment, which includes dynamic and static objects, e.g., drivable areas, buildings, pedestrian crossings. Camera, lidar, radar and ultrasonic sensors are the most commonly used modalities for this task. A detailed comparison of exteroceptive sensors is given in Table \ref{table_sensors_comp}. 

\subsubsection{Monocular Cameras}

Cameras can sense color and are passive, i.e. {they do} not emit any signal for measurements. Sensing color is extremely important for tasks such as traffic light recognition.  Furthermore, 2D computer vision is an established field with remarkable state-of-the-art algorithms. Moreover, a passive sensor does not interfere with other systems since it does not emit any signals.  However, cameras have certain shortcomings. Illumination conditions affect their performance drastically, and depth information is difficult to obtain from a single camera. There are promising studies \cite{saxena2006learning} to improve monocular camera based depth perception, but modalities that are not negatively affected by illumination and weather conditions are still necessary for dynamic driving tasks. Other camera types gaining interest for ADS include flash cameras\cite{jang2017design}, thermal cameras \cite{fries2015autonomous, ha2017mfnet}, and event cameras\cite{maqueda2018event}.

\subsubsection{Omnidirectional Camera}

\begin{figure}
	\centering
	\includegraphics[width=0.9\columnwidth]{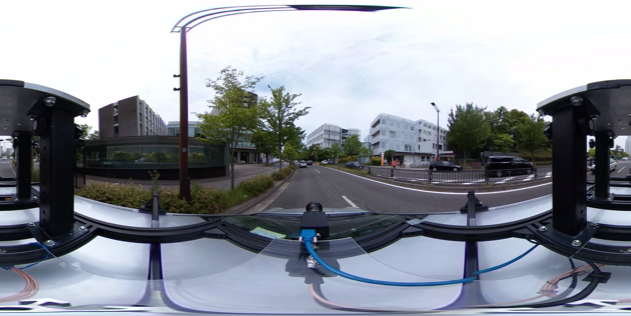}
	\caption{Ricoh Tetha V panoramic images collected using our data collection platform, in Nagoya University campus. Note some distortion still remains on the periphery of the image.}\label{fig:ricoh}
\end{figure}

For 360$^\circ$ 2D vision, omnidirectional cameras are used as an alternative to camera arrays. They have seen widespread use, with increasingly compact and high performance hardware being constantly released. Panoramic view is particularly desirable for applications such as navigation, localization and mapping\cite{Janai2017-my}. An example panoramic image is shown in Figure \ref{fig:ricoh}.

\begin{figure}
	\centering
	\includegraphics[width=0.9\columnwidth]{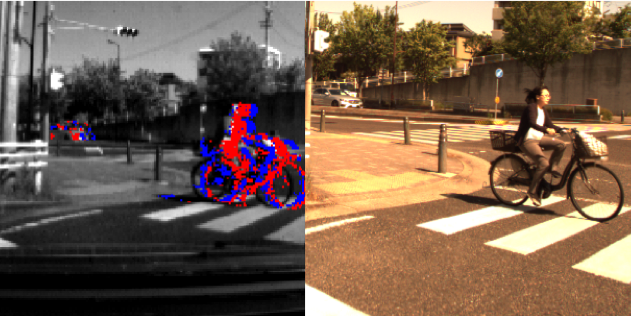}
	\caption{DAVIS240 events, overlayed on the image (left) and corresponding RBG image from a different camera (right), collected by our data collection platform, at a road crossing near Nagoya University. The motion of the cyclist and vehicle causes brightness changes which trigger events.}\label{fig:event}
\end{figure}

\subsubsection{Event Cameras}
Event cameras are among the newer sensing modalities that have seen use in ADS \cite{binas2017ddd17}. Event cameras record data asynchronously for individual pixels with respect to visual stimulus. The output is therefore an irregular sequence of data points, or \emph{events} triggered by changes in brightness. The response time is in the order of microseconds\cite{Lichtsteiner2008-ro}. The main limitation of current event cameras is pixel size and image resolution. For example, the DAVIS40 image shown in \Cref{fig:event} has a pixel size of $18.5 \times 18.5$ $\mu$m and a resolution of $240\times 180$. Recently, a driving dataset with event camera data has been published \cite{binas2017ddd17}.

\subsubsection{Radar}

Radar, lidar and ultrasonic sensors are very useful in covering the shortcomings of cameras. Depth information, i.e. distance to objects, can be measured effectively to retrieve 3D information with these sensors, and they are not affected by illumination conditions. However, they are active sensors. Radars emit radio waves that bounce back from objects and measure the time of each bounce. Emissions from active sensors can interfere with other systems. Radar is a well-established technology that is both lightweight and cost-effective. For example, radars can fit inside side-mirrors. Radars are cheaper and can detect objects at longer distances than lidars, but the latter are more accurate.

\subsubsection{Lidar}

\begin{figure}
	\centering\includegraphics[width=0.9\columnwidth]{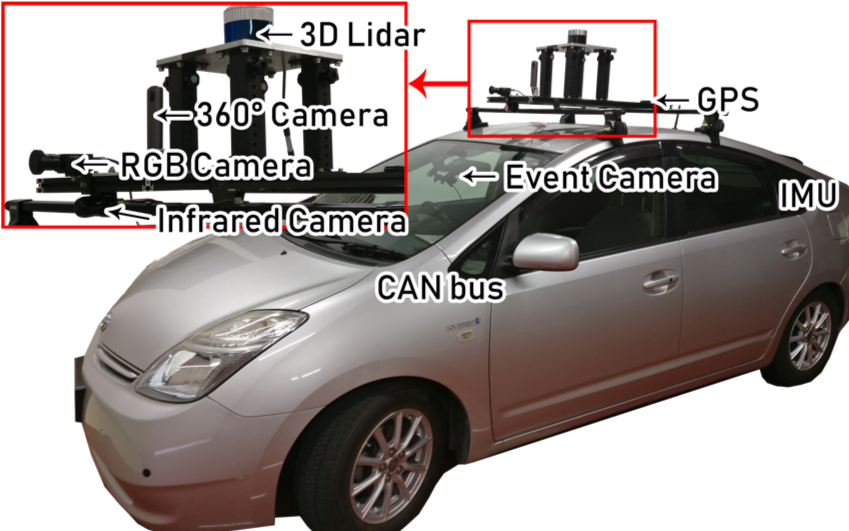}
	\caption{The ADS equipped Prius of Nagoya University. We have used this vehicle to perform core automated driving operations.}
	\label{fig_nagoya_prius}
\end{figure}

Lidar operates with a similar principle that of radar but it emits infrared light waves instead of radio waves. It has much higher accuracy than radar under 200 meters. Weather conditions such as fog or snow have a negative impact on the performance of lidar. Another aspect is the sensor size: smaller sensors are preferred on the vehicle because of limited space and aerodynamic restraints and lidars are generally larger than radars.

In \cite{schoettle2017sensor}, human sensing performance is compared to ADS. One of the key findings of this study is that even though human drivers are still better at reasoning in general, the perception capability of ADSs with sensor-fusion can exceed humans, especially in degraded conditions such as insufficient illumination.

\subsubsection{Proprioceptive sensors}

Proprioceptive sensing is another crucial category. Vehicle states such as speed, acceleration and yaw must be continuously measured in order to operate the platform safely with feedback. Almost all of the modern production cars are equipped with proprioceptive sensors. Wheel encoders are mainly used for odometry, Inertial Measurement Units (IMU) are employed for monitoring the velocity and position changes, tachometers are utilized for measuring speed and altimeters for altitude. These signals can be accessed through the CAN protocol of modern cars.

Besides sensors, an ADS needs actuators to manipulate the vehicle and advanced computational units for processing and storing sensor data.

\subsubsection{Full size cars}

There are numerous instrumented vehicles introduced by different research groups, such as Stanford's Junior \cite{levinson2011towards}, which employs an array of sensors with different modalities for perceiving external and internal variables. Boss won the DARPA Urban Challenge with an abundance of sensors \cite{urmson2008autonomous}. RobotCar \cite{maddern20171} is a cheaper research platform aimed for data collection. In addition, different levels of driving automation have been introduced by the industry; Tesla's Autopilot \cite{model2016autopilot} and Google's self driving car \cite{urmson2016google} are some examples. Bertha \cite{ziegler2014making} is developed by Daimler and has 4 $120^{\circ}$ short-range radars, two long-range range radar on the sides, stereo camera, wide angle-monocular color camera on the dashboard, another wide-angle camera for the back. Our vehicle is shown in \Cref{fig_nagoya_prius}.  A detailed comparison of sensor setups of 10 different full-size ADSs is given in Table \ref{table_sensors}.

%
%
\begin{table}[!t]
	\caption{Onboard sensor setup of ADS equipped vehicles}
	\label{table_sensors} 
	{\begin{tabularx}{\columnwidth}{c @{\extracolsep{\fill}} cccccc}
			\specialrule{.1em}{.05em}{.05em} 
			\B \T
			Platform &  \thead{\# $360^{\circ}$ rotating \\ lidars} & \thead{\# stationary \\ lidars} & \# Radars & \# Cameras \\ 
			\hline
			\textbf{Ours} & 1 & - & - & 4 \\
			Boss \cite{urmson2008autonomous} & 1 & 9 & 5 & 2 \\ 
			Junior \cite{levinson2011towards}  & 1  & 2 & 6 & 4 \\ 
			BRAiVE \cite{broggi2013extensive} & - & 5 & 1 & 10 \\ 
			RobotCar \cite{maddern20171} & - & 3 & - & 4 \\ 
			\hline
			Google car (prius) \cite{guizzo2011google} & 1 & - & 4 & 1 \\ 
			Uber car (XC90) \cite{somerville2018uber} & 1 & - & 10 & 7 \\ 
			Uber car (Fusion) \cite{somerville2018uber} & 1 & 7 & 7 & 20 \\ 
			Bertha \cite{ziegler2014making} & - & - & 6 & 3\\ 
			Apollo Auto \cite{apolloauto}  & 1 & 3 & 2 & 2 \\ 
			
			\specialrule{.1em}{.05em}{.05em} 
	\end{tabularx}}{}
\end{table}

\begin{table*}[!t]
	\caption{Localization techniques}
	\label{table_localization}
	{\begin{tabular*}{\textwidth}{l @{\extracolsep{\fill}} cccccc}
			\specialrule{.1em}{.05em}{.05em} 
			\B \T
			Methods & Robustness & Cost & Accuracy & Size & Computational requirements &  Related works   \\ 
			\hline
			Absolute positioning sensors &  low & low & low & small  & lowest & \cite{al2012hybrid} \\
			Odometry/dead reckoning & low  & low & low & smallest & low & \cite{chong1997accurate}  \\
			GPS-IMU fusion & medium  & medium & low &  small & low &  \cite{urmson2004high} \\
			SLAM &  medium-high & medium & high & large  & very high & \cite{bailey2006simultaneous}  \\
			\textbf{A priori Map-based} & & & & & &\\
			\hline
			\hspace{0.5cm} Landmark search & high & medium & high & large & medium & \cite{hata2014road},\cite{ort2018autonomous}\\
			\hspace{0.5cm} Point cloud matching & highest & highest & highest & largest & high & \cite{levinson2010robust}, \cite{takeuchi20063}\\
			\specialrule{.1em}{.05em}{.05em} 
		\end{tabular*}}{}
	\end{table*}

\subsubsection{Large vehicles and trailers}


 
Earliest intelligent trucks were developed for the PATH program in California \cite{shladover2007path}, which utilized magnetic markers on the road. Fuel economy is an essential topic in freight transportation and methods such as platooning has been developed for this purpose. Platooning is a well-studied phenomenon; it reduces drag and therefore fuel consumption \cite{alam2015heavy}. In semi-autonomous truck platooning, the lead truck is driven by a human driver, and several automated trucks follow it; forming a semi-autonomous road-train as defined in \cite{bergenhem2012overview}. Sartre European Union project \cite{chan2016sartre} introduced such a system that satisfies three core conditions: using the already existing public road network, sharing the traffic with non-automated vehicles and not modifying the road infrastructure. A platoon consisting of three automated trucks was formed in \cite{alam2015heavy} and significant fuel savings were reported.  


Tractor-trailer setup poses an additional challenge for automated freight transport. Conventional control methods such as feedback linearization \cite{khalaji2014robust} and fuzzy control \cite{cheng2017backward} were used for path tracking without considering the jackknifing constraint. The possibility of jackknifing, the collision of the truck and the trailer with each other, increases the difficulty of the task \cite{hejase2018constrained}. A control safety governor design was proposed in \cite{hejase2018constrained} to prevent jackknifing while reversing.

\section{Localization and mapping}\label{sec_localization}

Localization is the task of finding ego-position relative to a reference frame in an environment \cite{kuutti2018survey}, and it is fundamental to any mobile robot. It is especially crucial for ADSs \cite{bresson2017simultaneous}; the vehicle must use the correct lane and position itself in it accurately. Furthermore, localization is an elemental requirement for global navigation.  

The reminder of this section details the three most common approaches that use solely on-board sensors: Global Positioning System and Inertial Measurement Unit (GPS-IMU) fusion, Simultaneous Localization And Mapping (SLAM), and state-of-the-art a priori map-based localization. Readers are referred to \cite{kuutti2018survey} for a broader localization overview. A comparison of localization methods is given in Table \ref{table_localization}.

%
%


\subsection{GPS-IMU fusion}

The main principle of GPS-IMU fusion is correcting accumulated errors of dead reckoning in intervals with absolute position readings \cite{zhang2012sensor}. In a GPS-IMU system, changes in position and orientation are measured by IMU, and this information is processed for localizing the vehicle with dead reckoning. There is a significant drawback of IMU, and in general dead reckoning: errors accumulate with time and they often lead to failure in long-term operations \cite{kao1991integration}. With the integration of GPS readings, the accumulated errors of the IMU can be corrected in intervals.  

GPS-IMU systems by themselves cannot be used for vehicle localization as they do not meet the performance criteria \cite{levinson2007map}. In the 2004 DARPA Grand Challenge, the red team from Carnegie Mellon University \cite{urmson2004high} failed the race because of a GPS error. The accuracy required for urban automated driving is too high for the current GPS-IMU systems used in production cars. Moreover, in dense urban environments, the accuracy drops further, and the GPS stops functioning from time to time because of tunnels \cite{zhang2012sensor} and high buildings.  

Even though GPS-IMU systems by themselves do not meet the performance requirements and can only be utilized for high-level route planning, they are used for initial pose estimation in tandem with lidar and other sensors in state-of-the-art localization systems \cite{levinson2007map}.

\subsection{Simultaneous localization and mapping}

Simultaneous localization and mapping (SLAM) is the act of online map making and localizing the vehicle in it at the same time.  A priori information about the environment is not required in SLAM. It is a common practice in robotics, especially in indoor environments. However, due to the high computational requirements and environmental challenges, running SLAM algorithms outdoors, which is the operational domain of ADSs, is less efficient than localization with a pre-built map \cite{ranganathan2013light}. 



Team MIT used a SLAM approach in DARPA urban challenge  \cite{leonard2008perception} and finished it in the 4th place. Whereas, the winner, Carnegie Mellon`s Boss \cite{urmson2008autonomous} and the runner-up, Stanford`s Junior \cite{levinson2011towards}, both utilized a priori information. In spite of not having the same level of accuracy and efficiency, SLAM techniques have one major advantage over a priori methods: they can work anywhere. 

SLAM based methods have the potential to replace a priori techniques if their performances can be increased further \cite{van2018autonomous}. We refer the readers to \cite{bresson2017simultaneous} for a detailed SLAM survey in the intelligent vehicle domain.    




\subsection{A priori map-based localization}


The core idea of a priori map-based localization techniques is matching: localization is achieved through the comparison of online readings to the information on a detailed pre-built map and finding the location of the best possible match \cite{levinson2007map}. Often an initial pose estimation, for example with a GPS, is used at the beginning of the matching process. There are various approaches to map building and preferred modalities.   

%

Changes in the environment affect the performance of map-based methods negatively. This effect is prevalent especially in rural areas where past information of the map can deviate from the actual environment because of changes in roadside vegetation and constructions \cite{akai2017robust}. Moreover, this method requires an additional step of map making.

There are two different map-based approaches; landmark search and matching.

\subsubsection{Landmark search}

Landmark search is computationally less expensive in comparison to point cloud matching. It is a robust localization technique as long as a sufficient amount of landmarks exists. In an urban environment, poles, curbs, signs and road markers can be used as landmarks. 

A road marking detection method using lidar and Monte Carlo Localization (MCL) was used in \cite{hata2014road}. In this method, road markers and curbs were matched to a 3D map to find the location of the vehicle. A vision based road marking detection method was introduced in \cite{suhr2017sensor}. Road markings detected by a single front camera were compared and matched to a low-volume digital marker map with global coordinates. Then, a particle filter was employed to update the position and heading of the vehicle with the detected road markings and GPS-IMU output. A road marking detection based localization technique using; two cameras directed towards the ground, GPS-IMU dead reckoning, odometry, and a precise marker location map was proposed in \cite{gruyer2016accurate}. Another vision based method with a single camera and geo-referenced traffic signs was presented in \cite{qu2015vehicle}.

This approach has one major disadvantage; landmark dependency makes the system prone to fail where landmark amount is insufficient.  



\begin{figure}[!t]
	\centering\includegraphics[width=0.8\columnwidth]{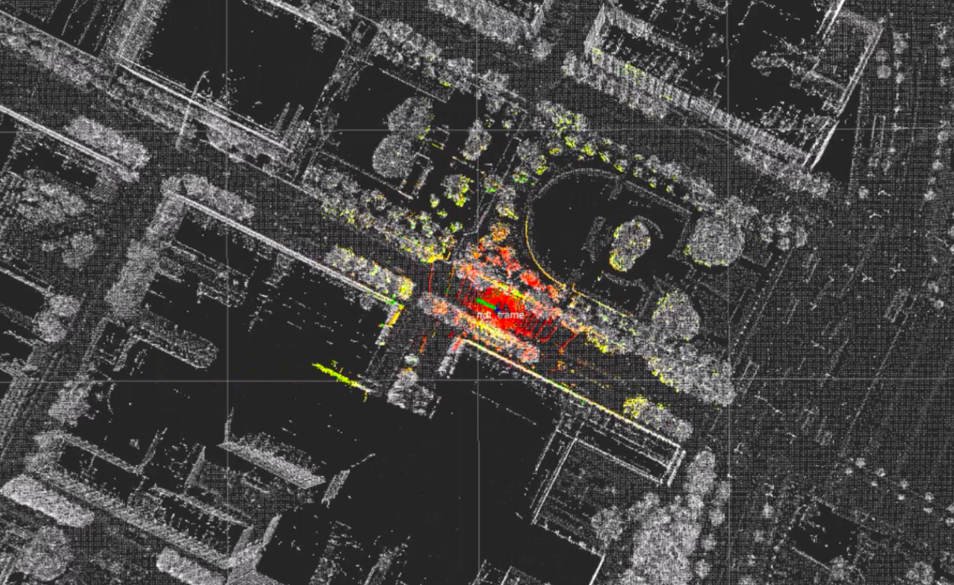}
	\caption{We used NDT matching \cite{takeuchi20063,magnusson2009} to localize our vehicle in the Nagoya University campus. White points belong to the offline pre-built map and the colored ones were obtained from online scans. The objective is to find the best match between colored points and white points, thus localizing the vehicle.} 
	\label{fig_ndt}
\end{figure}

\subsubsection{Point cloud matching}

\begin{figure}
	\centering\includegraphics[width=0.8\columnwidth]{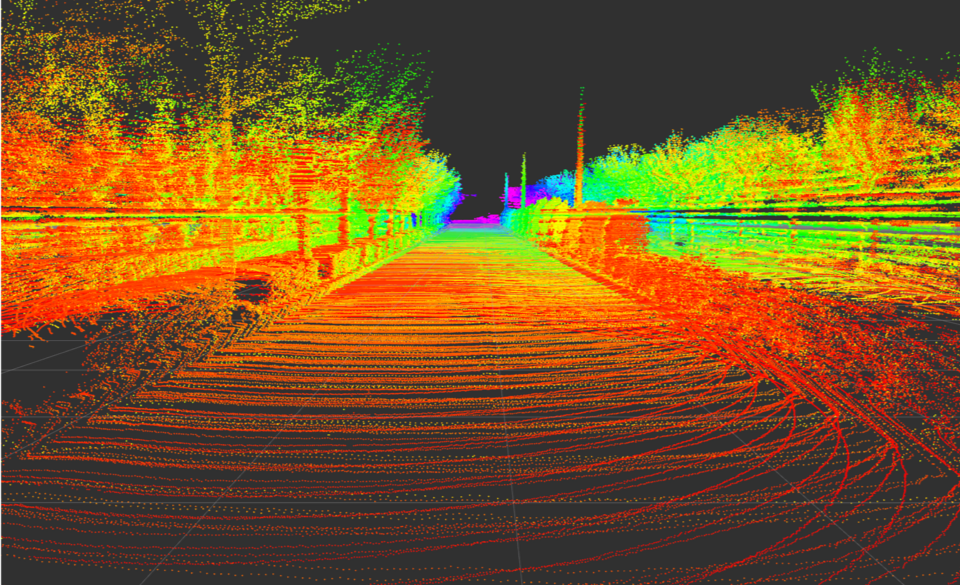}
	\caption{Creating a 3D pointcloud map with congregation of scans. We used Autoware \cite{kato2015open} for mapping. }
	\label{fig_map_making}
\end{figure}

The state-of-the-art localization systems use multi-modal point cloud matching based approaches. In summary, the online-scanned point cloud, which covers a smaller area, is translated and rotated around its center iteratively to be compared against the larger a priori point cloud map. The position and orientation that gives the best match between the two point clouds give the localized position of the sensor relative to the map. For initial pose estimation, GPS is used commonly along dead reckoning. We used this approach to localize our vehicle. The matching process is shown in Figure \ref{fig_ndt} and the map-making in Figure \ref{fig_map_making}.      

In the seminal work of \cite{levinson2007map}, a point cloud map collected with lidar was used to augment inertial navigation and localization. A particle filter maintained a three-dimensional vector of 2D coordinates and the yaw angle. A multi-modal approach with probabilistic maps was utilized in \cite{levinson2010robust} to achieve localization in urban environments with less than 10 cm RMS error. Instead of comparing two point clouds point by point and discarding the mismatched reads, the variance of all observed data was modeled and used for the matching task.  A matching algorithm for lidar scans using multi-resolution Gaussian Mixture Maps (GMM) was proposed in \cite{wolcott2015fast}. Iterative Closest Point (ICP) was compared against Normal Distribution Transform (NDT) in \cite{magnusson2009evaluation,magnusson2009}. In NDT, accumulated sensor readings are transformed into a grid that is represented by the mean and covariance obtained from the scanned points that fall into its' cells/voxels. NDT proved to be more robust than point-to-point ICP matching. An improved version of 3D NDT matching was proposed in \cite{takeuchi20063}, and \cite{akai2017robust} augmented NDT with road marker matching. An NDT-based Monte Carlo Localization (MCL) method that utilizes an offline static map and a constantly updated short-term map was developed by  \cite{valencia2014localization}. In this method, NDT occupancy grid was used for the short-term map and it was utilized only when and where the static map failed to give sufficient explanations. 

Map-making and maintaining is time and resource consuming. Therefore some researchers such as \cite{ort2018autonomous} argue that methods with a priori maps are not feasible given the size of road networks and rapid changes. 

\subsubsection{2D to 3D matching}

Matching online 2D readings to a 3D a priori map is an emerging technology. This approach requires only a camera on the ADS equipped vehicle instead of the more expensive lidar. The a priori map still needs to be created with a lidar.

A monocular camera was used to localize the vehicle in a point cloud map in \cite{wolcott2014visual}. With an initial pose estimation, 2D synthetic images were created from the offline 3D point cloud map and they were compared with normalized mutual information to the online images received from the camera. This method increases the computational load of the localization task. Another vision matching algorithm was introduced in \cite{mcmanus2013distraction} where a stereo camera setup was utilized to compare online readings to synthetic depth images generated from 3D prior.  

Camera based localization approaches could become popular in the future as the hardware requirement is cheaper than lidar based systems.

\section{Perception}\label{sec:perception}

Perceiving the surrounding environment and extracting information which may be critical for safe navigation is a critical objective for ADS. A variety of tasks, using different sensing modalities, fall under the category of perception. 
Building on decades of computer vision research, cameras are the most commonly used sensor for perception, with 3D vision becoming a strong alternative/supplement.  

The reminder of this section is divided into core perception tasks. We discuss image-based object detection in \Cref{sec:image_detection}, semantic segmentation in \Cref{sec:semantic_segmentation}, 3D object detection in \Cref{sec:3d_detection}, road and lane detection in \Cref{sec:road_lane_detection} and object tracking in \Cref{sec:tracking}.


\subsection{Detection}

\subsubsection{Image-based Object Detection}\label{sec:image_detection}

\begin{table}
	\centering
	\caption{Comparison of 2D bounding box estimation architectures on the test set of ImageNet1K, ordered by Top 5\% error. Number of parameters (Num. Params) and number of layers (Num. Layers), hints at the computational cost of the algorithm.}\label{tab:image_results}
	{\footnotesize
		\begin{tabular}{c c c c c c}
			\specialrule{.1em}{.05em}{.05em} \T
			Architecture	&	Num. Params 	&	Num. 	&	ImageNet1K	\\ \B
			& ($\times 10^6$) & Layers & Top 5 Error \% \\
			\hline \T
			Incept.ResNet v2\cite{Szegedy2016-dg}	 &	30	&	95	&	4.9	\\
			Inception v4\cite{Szegedy2016-dg}&		41	&	75	&	5	\\
			ResNet101\cite{He2016-db}	&	45	&	100	&	6.05	\\
			DenseNet201\cite{Huang2016-vb}	&	18	&	200	&	6.34\\
			YOLOv3-608\cite{Redmon2018-ar} &		63	&	53+1	&	6.2	\\
			ResNet50\cite{He2016-db}	&	26	&	49	&	6.7	\\
			GoogLeNet\cite{Szegedy2015-kh}	&	6	&	22	&	6.7	\\
			VGGNet16\cite{Simonyan2015-nf}	&	134	&	13+2	&	6.8	\\
			AlexNet\cite{krizhevsky2012imagenet}	&	57	&	5+2	&	15.3 \\
			\specialrule{.1em}{.05em}{.05em}
	\end{tabular}}
\end{table}

Object detection refers to identifying the location and size of objects of interest. Both static objects, from traffic lights and signs to road crossings, and dynamic objects such as other vehicles, pedestrians or cyclists are of concern to ADSs. Generalized object detection has a long-standing history as a central problem in computer vision, where the goal is to determine if objects of specific classes are present in an image, then to determine their size via a rectangular bounding box. This section mainly discusses state-of-the-art object detection methods, as they represent the starting point of several other tasks in an ADS pipe, such as object tracking and scene understanding.

Object recognition research started more than 50 years ago, but only recently, in the late 1990s and early 2000s, has algorithm performance reached a level of relevance for driving automation. In 2012, the deep convolutional neural network (DCNN) AlexNet\cite{krizhevsky2012imagenet} shattered the ImageNet image recognition challenge\cite{Deng2009-yh}. This resulted in a near complete shift of focus to supervised learning and in particular deep learning for object detection. There exists a number of extensive surveys on general image-based object detection\cite{Andreopoulos2013-bc, Zhao2018-su, Liu2018-jm}. Here, the focus is on the state-of-the-art methods that could be applied to ADS.



While state-of-the-art methods all rely on DCNNs, there currently exist a clear distinction between them:
\begin{enumerate}
\item Single stage detection frameworks use a single network to produce object detection locations and class prediction simultaneously.
\item Region proposal detection frameworks use two distinct stages, where general regions of interest are first proposed, then categorized by separate classifier networks. \\
\end{enumerate}

Region proposal methods are currently leading detection benchmarks, but at the cost requiring high computation power, and generally being difficult to implement, train and fine-tune. Meanwhile, single stage detection algorithms tend to have fast inference time and low memory cost, which is well-suited for real-time driving automation. YOLO (You Only Look Once)\cite{Redmon2016-fo} is a popular single stage detector, which has been improved continuously\cite{Redmon2017-fo, Redmon2018-ar}. Their network uses a DCNN to extract image features on a coarse grid, significantly reducing the resolution of the input image. A fully-connected neural network then predicts class probabilities and bounding box parameters for each grid cell and class. This design makes YOLO very fast, the full model operating at 45 FPS and a smaller model operating at 155 FPS for a small accuracy trade-off. More recent versions of this method, YOLOv2, YOLO9000\cite{Redmon2017-fo} and YOLOv3\cite{Redmon2018-ar} briefly took over the PASCAL VOC and MS COCO benchmarks while maintaining low computation and memory cost. Another widely used algorithm, even faster than YOLO, is the Single Shot Detector (SSD)\cite{Liu2015-hf}, which uses standard DCNN architectures such as VGG\cite{Simonyan2015-nf} to achieve competitive results on public benchmarks. SSD performs detection on a coarse grid similar to YOLO, but also uses higher resolution features obtained early in the DCNN to improve detection and localization of small objects.

Considering both accuracy and computational cost is essential for detection in ADS; the detection needs to be reliable, but also operate better than real-time, to allow as much time as possible for the planning and control modules to react to those objects. As such, single stage detectors are often the detection algorithms of choice for ADSs. However, as shown in \Cref{tab:image_results}, region proposal networks (RPN), used in two-stage detection frameworks, have proven to be unmatched in terms of object recognition and localization accuracy, and computational cost has improved greatly in recent years. They are also better suited for other tasks related to detection, such as semantic segmentation as discussed in \Cref{sec:semantic_segmentation}. Through transfer learning, RPNs achieving multiple perception tasks simultaneously are become increasingly feasible for online applications\cite{He2017-zt}. RPNs can replace single stage detection networks for ADS applications in the near future.

\begin{figure*}
	\centering
	\includegraphics[width=0.8\textwidth]{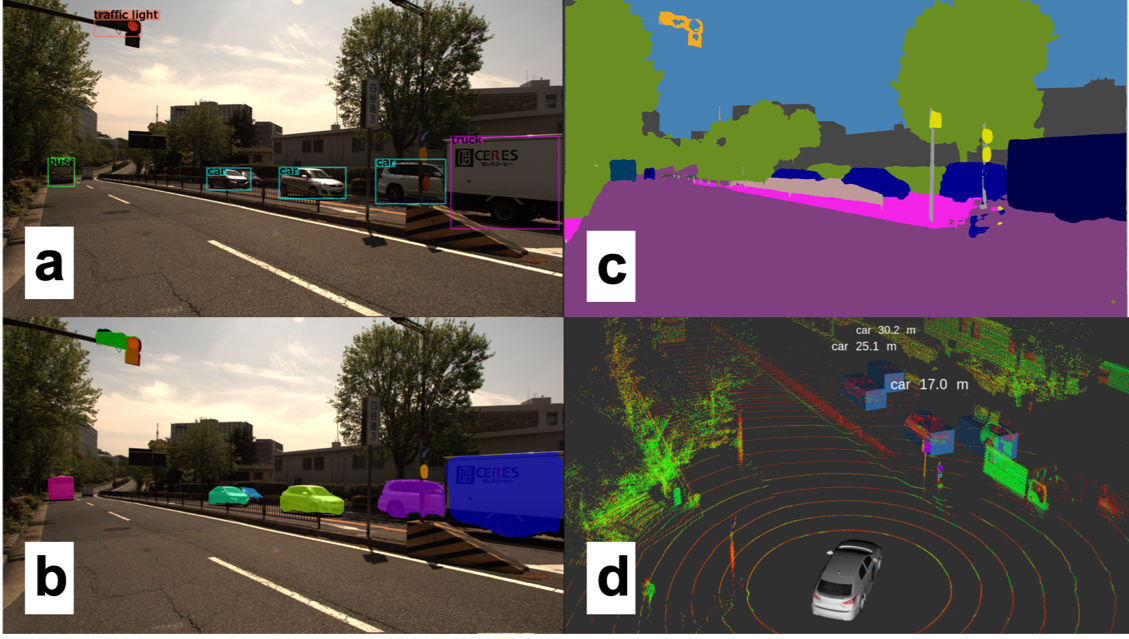}
	\caption{An urban scene near Nagoya University, with camera and lidar data collected by our experimental vehicle and object detection outputs from state-of-the-art perception algorithms.  (a) A front facing camera's view, with bounding box results from YOLOv3\cite{Redmon2018-ar} and (b) instance segmentation results from MaskRCNN\cite{He2017-zt}. (c) Semantic segmentation masks produced by DeepLabv3\cite{Chen2018-fo}. (d) The 3D Lidar data with object detection results from SECOND\cite{Yan2018-rl}. Amongst the four, only the 3D perception algorithm outputs range to detected objects.}\label{fig:lidar-image}
\end{figure*}

\textbf{Omnidirectional and event camera-based perception:}
360 degree vision, or at least panoramic vision, is necessary for higher levels of automation. This can be achieved through camera arrays, though precise extrinsic calibration between each camera is then necessary to make image stitching possible. Alternatively, omnidirectional cameras can be used, or a smaller array of cameras with very wide angle \emph{fisheye} lenses. These are however difficult to intrinsically calibrate; the spherical images are highly distorted and the camera model used must account for mirror reflections or fisheye lens distortions, depending on the camera model producing the panoramic images\cite{Geyer2000-yz, Scaramuzza2006-dc}. The accuracy of the model and calibration dictates the quality of undistorted images produced, on which the aforementioned 2D vision algorithms are used. An example of fisheye lenses producing two spherical images then combined into one panoramic image is shown in \Cref{fig:ricoh}. Some distortions inevitably remain, but despite these challenges in calibration, omnidirectional cameras have been used for many applications such as SLAM\cite{Scaramuzza2008-xs} and 3D reconstruction \cite{Schonbein2014-zi}.

Event cameras are a fairly new modality which output asynchronous \emph{events} usually caused by movement in the observed scene, as shown in \Cref{fig:event}. This makes the sensing modality interesting for dynamic object detection. The other appealing factor is their response time on the order of microseconds\cite{Lichtsteiner2008-ro}, as frame rate is a significant limitation for high-speed driving. The sensor resolution remains an issue, but new models are rapidly improving. They have been used for a variety of applications closely related to ADS. A recent survey outlines progress in pose estimation and SLAM, visual-inertial odometry and 3D reconstruction, as well as other applications\cite{Gallego2019-sg}. Most notably, a dataset for end-to-end driving with event cameras was recently published, with preliminary experiments showing that the output of an event camera can, to some extent, be used to predict car steering angle \cite{binas2017ddd17}.

\textbf{Poor Illumination and Changing Appearance:}
The main drawback with using camera is that changes in lighting conditions can significantly affect their performance. Low light conditions are inherently difficult to deal with, while changes in illumination due to shifting shadows, intemperate weather, or seasonal changes, can cause algorithms to fail, in particular supervised learning methods. For example, snow drastically alters the appearance of scenes and hides potentially key features such as lane markings. An easy alternative is to use an alternate sensing modalities for perception, but lidar also has difficulties with some weather conditions like fog and snow\cite{Rasshofer2005-ay}, and radars lack the necessary resolution for many perception tasks\cite{wei2013towards}. A sensor fusion strategy is often employed to avoid any single point of failure\cite{Radecki2016-ul}.

Thermal imaging through infrared sensors are also used for object detection in low light conditions, which is particularly effective for pedestrian detection\cite{Hurney2015-jl}. Camera-only methods which attempt to deal with dynamic lighting conditions directly have also been developed. Both attempting to extract lighting invariant features\cite{Carlevaris-Bianco2014-sv} and assessing the quality of features\cite{Peretroukhin2017-kq} have been proposed. Pre-processed, \emph{illumination invariant} images have applied to ADS\cite{Maddern2014-tt} and were shown to improve localization, mapping and scene classification capabilities over long periods of time. Still, dealing with the unpredictable conditions brought forth by inadequate or changing illumination remains a central challenge preventing the widespread implementation of ADS.

\begin{figure*}
	\centering
	\includegraphics[width=0.9\textwidth]{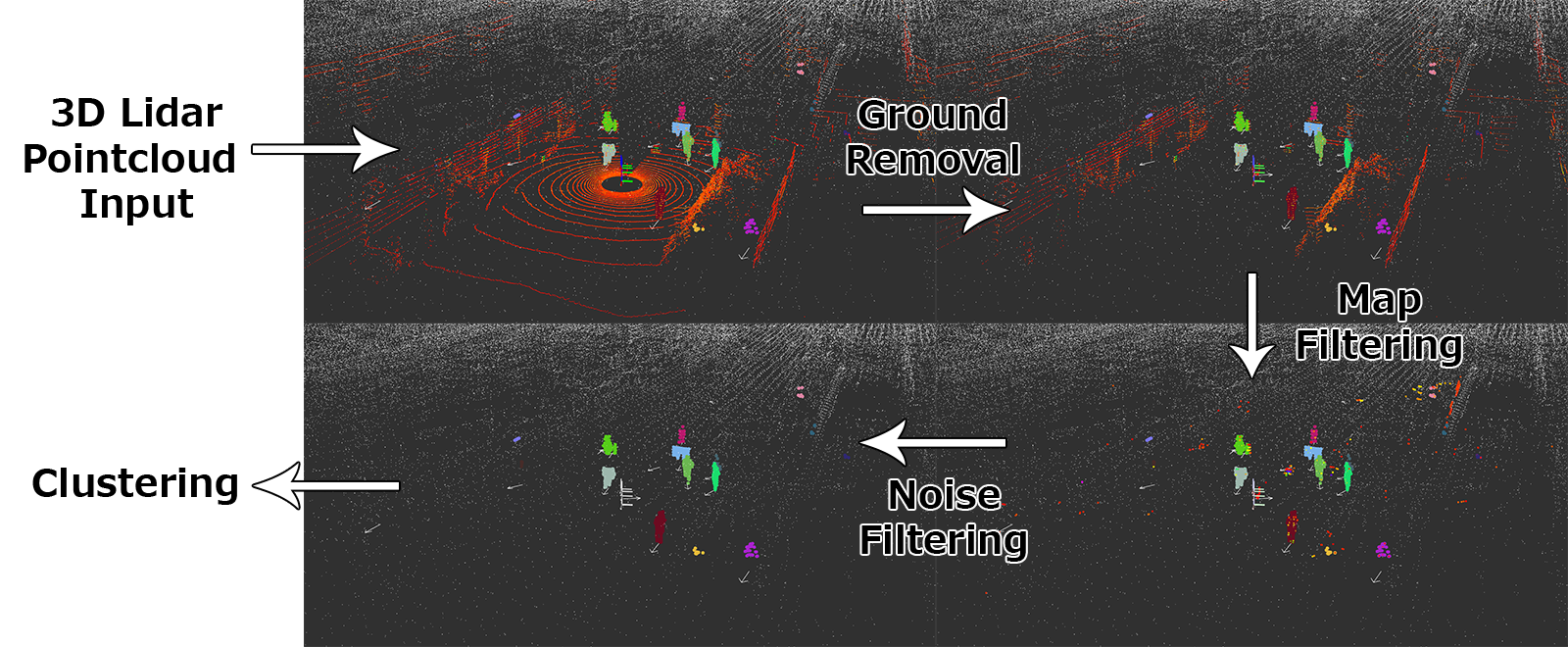}
	\caption{Outline of a traditional method for object detection from 3D pointcloud data. Various filtering and data reduction methods are used first, followed by clustering. The resulting clusters are shown by the different colored points in the 3D lidar data of pedestrians collected by our data collection platform.}\label{fig:lidar-pipeline}
\end{figure*}

\subsubsection{Semantic Segmentation}\label{sec:semantic_segmentation}

Beyond image classification and object detection, computer vision research has also tackled the task of image segmentation. This consists of classifying each pixel of an image with a class label. This task is of particular importance to driving automation as some objects of interest are poorly defined by bounding boxes, in particular roads, traffic lines, sidewalks and buildings. A segmented scene in an urban area can be seen in \Cref{fig:lidar-image}. As opposed to semantic segmentation, which labels pixels based on a class, instance segmentation algorithms further separates instances of the same class, which is important in the context of driving automation. In other words, objects which may have different trajectories and behaviors must be differentiated from each other. We used the COCO dataset\cite{Lin2014-rh} to train the instance segmentation algorithm Mask R-CNN\cite{He2017-zt} with the sample result shown in \Cref{fig:lidar-image}.




Segmentation has recently become feasible for real-time applications. Generally, developments in this field progress in parallel with image-based object detection. The aforementioned Mask R-CNN\cite{He2017-zt} is a generalization of Faster R-CNN\cite{Ren2015-va}. The multi-task R-CNN network can achieve accurate bounding box estimation and instance segmentation simultaneously and can also be generalized to other tasks like pedestrian pose estimation with minimal domain knowledge. Running at 5 fps means it is approaching the area of real-time use for ADS. 

Unlike Mask-RCNN's architecture which is more akin to those used for object detection through its use of region proposal networks, segmentation networks usually employ a combination of convolutions for feature extraction. Those are followed by deconvolutions, also called transposed convolutions, to obtain pixel resolution labels\cite{Noh2015-ld, Ronneberger2015-ty}. Feature pyramid networks are also commonly used, for example in PSPNet \cite{Zhao2016-li}, which also introduced dilated convolutions for segmentation. This idea of sparse convolutions was then used to develop DeepLab\cite{Chen2016-dt}, with the most recent version being the current state-of-the-art for object segmentation\cite{Chen2018-fo}. We employed DeepLab with our ADS and a segmented frame is shown in \Cref{fig:lidar-image}.

While most segmentation networks are as of yet too slow and computationally expensive to be used in ADS, it is important to notice that many of these segmentations networks are initially trained for different tasks, such as bounding box estimation, then generalized to segmentation networks. Furthermore, these networks were shown to learn universal feature representations of images, and can be generalized for many tasks. This suggests the possibility that single, generalized perception networks may be able to tackle all perception tasks required for an ADS. 

\subsubsection{3D Object Detection}\label{sec:3d_detection}

Given their affordability, availability and widespread research, cameras are used by nearly all algorithms presented so far as the primary perception modality. However, cameras have limitations that are critical to ADS. Aside from illumination which was previously discussed, camera-based object detection occurs in the projected image space and therefore the scale of the scene is unknown. To make use of this information for dynamic driving tasks like obstacle avoidance, it is necessary to bridge the gap from 2D image-based detection to the 3D, metric space.  Depth estimation is therefore necessary, which is in fact possible with a single camera\cite{Ma2019-dp} though stereo or multi-view systems are more robust\cite{Cheng2018-le}. These algorithms necessarily need to solve an expensive image matching problem, which adds a significant amount of processing cost to an already complex perception pipeline. 


A relatively new sensing modality, the 3D lidar, offers an alternative for 3D perception. The 3D data collected inherently solves the scale problem, and since they have their own emission source, they are far less dependable on lighting condition, and less susceptible to intemperate weather. The sensing modality collects sparse 3D points representing the surfaces of the scene, as shown in \Cref{fig:lidar-pipeline}, which are challenging to use for object detection and classification. The appearance of objects change with range, and after some distance, very few data points per objects are available to detect an object. This poses some challenges for detection, but since the data is a direct representation of the world, it is more easily separable. Traditional methods often use euclidean clustering\cite{RusuDoctoralDissertation} or region-growing methods\cite{Wang2016-en} for grouping points into objects. This approach has been made much more robust through various filtering techniques, such as ground filtering\cite{Narksri2018-jj} and map-based filtering\cite{Lambert2018-ng}. We implemented a 3D object detection pipeline to get clustered objects from raw point cloud input. An example of this process is shown in \Cref{fig:lidar-pipeline}.

\begin{figure}
	\centering
	\includegraphics[width=0.8\columnwidth]{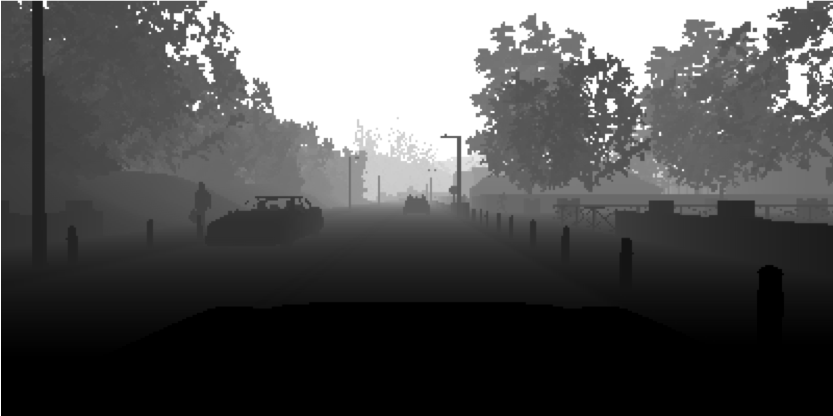}
	\caption{A depth image produced from synthetic lidar data, generated in the CARLA \cite{dosovitskiy2017carla} simulator.}\label{fig:depth_image}
\end{figure}

As with image-based methods, machine learning has also recently taken over 3D detection methods. These methods have also notably been applied to RGB-D\cite{Song2014-hn}, which produce similar, but colored, point clouds; with their limited range and unreliability outdoors, RGB-D have not been used for ADS applications. A 3D representation of point data, through a 3D occupancy grid called voxel grids, was first applied for object detection in RGB-D data\cite{Song2014-hn}. Shortly thereafter, a similar approach was used on point clouds created by lidars \cite{Wang-RSS-15}. Inspired by image-based methods, 3D CNNs are used, despite being computationally very expensive. 

The first convincing results for point cloud-only 3D bounding box estimation were produced by VoxelNet\cite{Zhou2017-xm}. Instead of hand-crafting input features computed during the discretization process, VoxelNet learned an encoding from raw point cloud data to voxel grid. Their voxel feature encoder (VFE) uses a fully connected neural network to convert the variable number of points in each occupied voxel to a feature vector of fixed size. The voxel grid encoded with feature vectors was then used as input to an aforementioned RPN for multi-class object detection. This work was then improved both in terms of accuracy and computational efficiency by SECOND\cite{Yan2018-rl} by exploiting the natural sparsity of lidar data. We employed SECOND and a sample result is shown in \Cref{fig:lidar-image}. Several algorithms have been produced recently, with accuracy constantly improving as shown in \Cref{tab:3d_kitti}, yet the computational complexity of 3D convolutions remains an issue for real-time use.

Another option for lidar-based perception is 2D projection of point cloud data. There are two main representations of point cloud data in 2D, the first being a so-called depth image shown in \Cref{fig:depth_image}, largely inspired by camera-based methods that perform 3D object detection through depth estimation\cite{Chen2015-uk} and methods that operate on RGB-D data \cite{Lin2013-gl}. The VeloFCN network \cite{Li-RSS-16} proposed to use single-channel depth image as input to a shallow, single-stage convolutional neural network which produced 3D vehicle proposals, with many other algorithms adopting this approach. Another use of depth image was shown for semantic classification of lidar points\cite{Liu2017-te}.

\begin{figure}
	\centering
	\includegraphics[width=0.8\columnwidth]{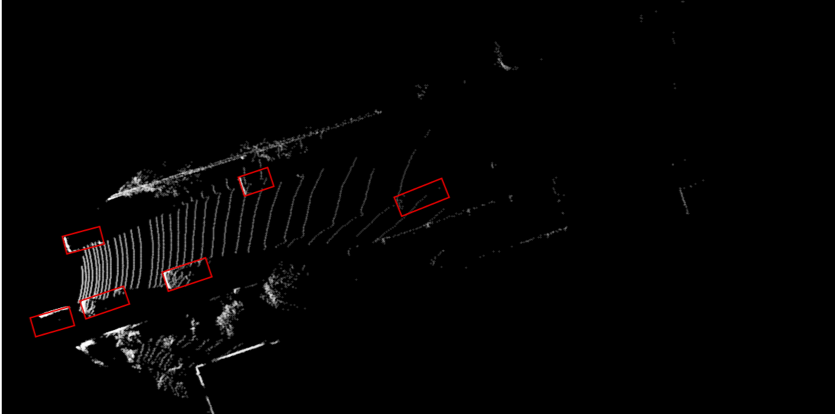}
	\caption{Bird's eye view perspective of 3D lidar data, a sample from the KITTI dataset\cite{geiger2012we}.}\label{fig:bv_image}
\end{figure}

The other 2D projection that has seen increasing popularity, in part due to the new KITTI benchmark, is projection to bird's eye view (BV) image. This is a top-view image of point clouds  as shown in \Cref{fig:bv_image}. Bird's eye view images discretize space purely in 2D, so lidar points which vary in height alone occlude each other. The MV3D algorithm\cite{Chen-CVPR-17} used camera images, depth images, as well as multi-channel BV images; each channel corresponding to a different range of heights, so as to minimize these occlusions. Several other works have reused camera-based algorithms and trained efficient networks for 3D object detection on 2D BV images\cite{Ren2018-ra, Ali2018-rz, Yang2018-xa, Feng2018-mw}. State-of-the-art algorithms are currently being evaluated on the KITTI dataset\cite{geiger2012we} and nuScenes dataset\cite{nuScene2018} as they offer labeled 3D scenes. \Cref{tab:3d_kitti} shows the leading methods on the KITTI benchmark, alongside detection times. 2D methods are far less computationally expensive, but recent methods that take point sparsity into account\cite{Yan2018-rl} are real-time viable and rapidly approaching the accuracy necessary for integration in ADSs.

\begin{table}
\centering
\caption{Average Precision (AP) in \% on the KITTI 3D object detection test set \emph{car} class, ordered based on \emph{moderate} category accuracy. These algorithms only use pointcloud data.}\label{tab:3d_kitti}
{\small
\begin{tabular}{ccccc}
\specialrule{.1em}{.05em}{.05em} \B \T
Algorithm &  T [s] & Easy & Moderate & Hard \\
\hline \T
PointRCNN\cite{Shi2018-qx}			&	0.10	&	85.9	&	75.8	&	68.3	\\
PointPillars\cite{Lang2018-sg}		&	0.02	&	79.1 	&	75.0	&	68.3	\\
SECOND\cite{Yan2018-rl}				&	0.04	&	83.1	&	73.7	&	66.2	\\
IPOD\cite{Yang2018-wx}				&	0.20	&	82.1	&	72.6	&	66.3	\\
F-PointNet\cite{Qi2017-yt}			&	0.17	&	81.2	&	70.4	&	62.2	\\
VoxelNet (Lidar)\cite{Zhou2017-xm}	&	0.23	&	77.5	&	65.1	&	57.7	\\ \B
MV3D (Lidar)\cite{Chen-CVPR-17}		&	0.24	&	66.8	&	52.8	&	51.3	\\  
\specialrule{.1em}{.05em}{.05em}
\end{tabular}}
\end{table}

\textbf{Radar}
Radar sensors have already been used for various perception applications, in various types of vehicles, with different models operating at complementary ranges. While not as accurate as the lidar, it can detect at object at high range and estimate their velocity \cite{leonard2008perception}. The lack of precision for estimating shape of objects is a major drawback when it is used in perception systems\cite{wei2013towards}, the resolution is simply too low. As such, it can be used for range estimation to large objects like vehicles, but it is challenging for pedestrians or static objects. Another issue is the very limited field of view of most radars, forcing a complicated array of radar sensors to cover the full field of view. Nevertheless, radar have seen widespread use as an ADAS component, for applications including proximity warning and adaptive cruise control\cite{Rasshofer2005-ay}. While radar and lidar are often seen as competing sensing modalities, they will likely be used in tandem in fully automated driving systems. Radars are very long range, have low cost and are robust to poor weather, while lidar offer precise object localization capabilities, as discussed in \Cref{sec_localization}. 

Another similar sensor to the radar are sonar devices, though their extremely short range of $<2m$ and poor angular resolution makes their use limited to very near obstacle detection \cite{Rasshofer2005-ay}.

\subsection{Object Tracking}\label{sec:tracking}

Object tracking is also often referred to as multiple object tracking (MOT)\cite{Luo2014-az} and detection and tracking of multiple objects (DATMO)\cite{Azim2012-pv}. For fully automated driving in complex and high speed scenarios, estimating location alone is insufficient. It is necessary to estimate dynamic objects' heading and velocity so that a motion model can be applied to track the object over time and predict future trajectory to avoid collisions. These trajectories must be estimated in the vehicle frame to be used by planning, so range information must be obtained through multiple camera systems, lidars or radar sensors. 3D lidars are often used for their precise range information and large field of view, allowing tracking over longer periods of time. To better cope with the limitations and uncertainties of different sensing modalities, a sensor fusion strategy is often use for tracking\cite{urmson2008autonomous}.   




\begin{figure}
\centering
\includegraphics[width=1\columnwidth]{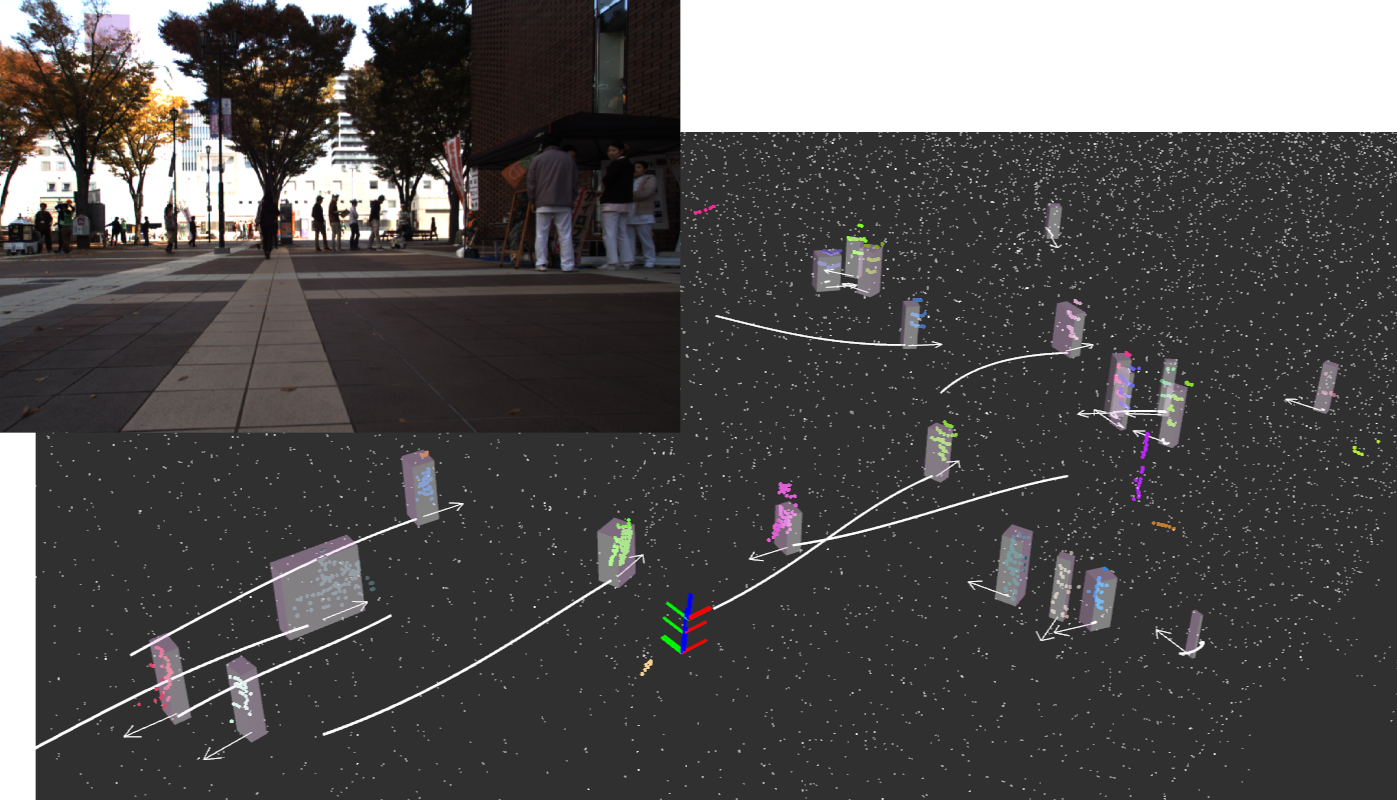}
\caption{A scene with several tracked pedestrians and cyclist with a basic particle filter on an urban road intersection. Past trajectories are shown in white with current heading and speed shown by the direction and magnitude of the arrow, sample collected by our data collection platform.}\label{fig:tracking}
\vspace{-0.25cm}
\end{figure}

Commonly used object trackers rely on simple data association techniques followed by traditional filtering methods. When objects are tracked in 3D space at high frame rate, nearest neighbor methods are often sufficient for establishing associations between objects. Image-based methods, however, need to establish some appearance model, which may consider the use of color histograms, gradients and other features such as KLT to evaluate the similarity \cite{Shi1994-wp}. Point cloud based methods may also use similarity metrics such as point density and Hausdorff distance\cite{Dubuisson1994-du, Lambert2018-ng}. Since association errors are always a possibility, multiple hypothesis tracking algorithms\cite{Hwang2016-cm} are often employed, which ensures tracking algorithms can recover from poor data association at any single time step. Using occupancy maps as a frame for all sensors to contribute to and then doing data association in that frame is common, especially when using multiple sensors\cite{Nguyen2012-ja}. To obtain smooth dynamics, the detection results are filtered by traditional Bayes filters. Kalman filtering is sufficient for simple linear models, while the extended and and unscented Kalman filters\cite{Ziegler2014-mf} are used to handle nonlinear dynamic models\cite{Ess2010-tp}. We implemented a basic particle filter based object-tracking algorithm, and an example of tracked pedestrians in contrasting camera and 3D lidar perspective is shown in \Cref{fig:tracking}. 

Physical models for the object being tracked are also often used for more robust tracking. In that case, non-parametric methods such as particle filters are used, and physical parameters such as the size of the object are tracked alongside dynamics\cite{Petrovskaya2009-xb}. More involved filtering methods such as Rao-Blackwellized particle filters have also been used to keep track of both dynamic variables and vehicle geometry variables for an L-shape vehicle model\cite{He2016-cn}. Various models have been proposed for vehicles and pedestrians, while some models generalize to any dynamic object\cite{Wang2015-ya}.

Finally, deep learning has also been applied to the problem of tracking, particularly for images. Tracking in monocular images was achieved in real-time through a CNN-based method\cite{Huval2015-uz, Held2016-kk}. Multi-task network which estimate object dynamics are also emerging\cite{Chowdhuri2017-ns} which further suggests that generalized networks tackling multiple perception tasks may be the future of ADS perception. 

\begin{figure*}
	\centering\includegraphics[width=1.9\columnwidth]{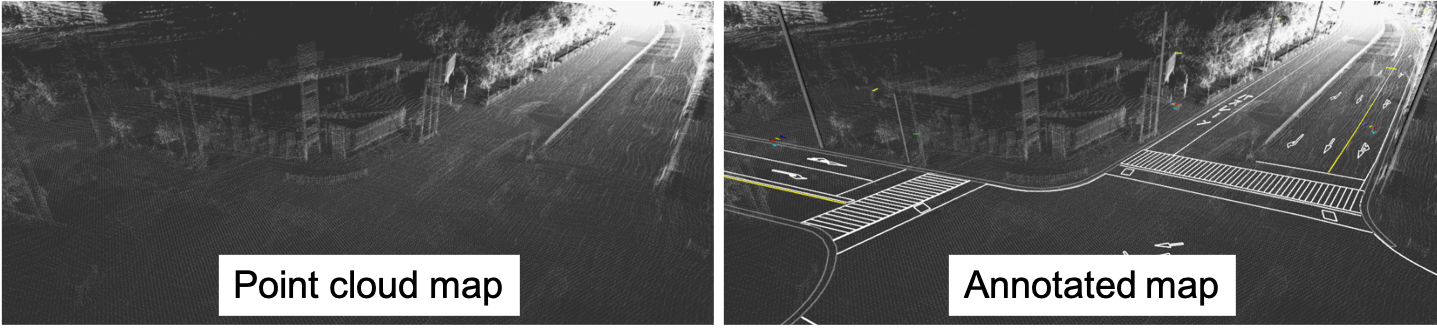}
	\caption{Annotating a 3D point cloud map with topological information. A large number of annotators were employed to build the map shown on the right-hand side. The point-cloud and annotated maps are available on \cite{autowareGit}.   }
	\label{fig_vector_map}
\end{figure*}

\subsection{Road and Lane Detection}\label{sec:road_lane_detection}

Bounding box estimation methods previously covered are useful for defining some objects of interest but are inadequate for continuous surfaces like roads. Determining the drivable surface is critical for ADSs and has been specifically researched as a subset of the detection problem. While drivable surfaces can be determined through semantic segmentation, automated vehicles need to understand road semantics to properly negotiate the road. An understanding of lanes, and how they are connected through merges and intersections remains a challenge from the perspective of perception. In this section, we provide an overview of current methods used for road and lane detection, and refer the reader to in-depth surveys of traditional methods\cite{McCall2006-gk} and the state-of-the-art methods\cite{hillel2014recent, Fernandez2017-xh}.


This problem is usually subdivided in several tasks, each unlocking some level of automation. The simplest is determining the drivable area from the perspective of the ego-vehicle. The road can then be divided into lanes, and the vehicles' host lane can be determined. Host lane estimation over a reasonable distance allows ADAS technology such as lane departure warning, lane keeping and adaptive cruise control \cite{McCall2006-gk, Labayrade2006-ta}.  Even more challenging is determining other lanes and their direction \cite{Yan_Jiang2010-aa}, and finally understanding complex semantics, as in their current and future direction, or merging and turning lanes \cite{urmson2008autonomous}. These ADAS or ADS technologies have different criteria both in terms of task, detection distance and reliability rates, but fully automated driving will require a complete, semantic understanding of road structures and the ability to detect several lanes at long ranges \cite{hillel2014recent}. Annotated maps as shown in  \Cref{fig_vector_map} are extremely useful for understanding lane semantics.

Current methods on road understanding typically first rely on exteroceptive data preprocessing. When cameras are used, this usually means performing image color corrections to normalize lighting conditions\cite{paton2015-is}. For lidar, several filtering methods can be used to reduce clutter in the data such as ground extraction\cite{Narksri2018-jj} or map-based filtering\cite{Lambert2018-ng}. For any sensing modality, identifying dynamic objects which conflicts with the static road scene is an important pre-processing step. Then, road and lane feature extraction is performed on the corrected data. Color statistics and intensity information\cite{Huang2009-ge}, gradient information\cite{Cheng2006-pc}, and various other filters have been used to detect lane markings. Similar methods have been used for road estimation, where the usual uniformity of roads and elevation gap at the edge allows for region growing methods to be applied\cite{Alvarez2007-th}. Stereo camera systems\cite{Danescu2009-qr}, as well as 3D lidars\cite{Huang2009-ge}, have been used determine the 3D structure of roads directly. More recently, machine learning-based methods which either fuse maps with vision\cite{Fernandez2017-xh} or use fully appearance-based segmentation\cite{long2015fully} have been used.

Once surfaces are estimated, model fitting is used to establish the continuity of the road and lanes. Geometric fitting through parametric models such as lines\cite{Borkar2009-wn} and splines\cite{Huang2009-ge} have been used, as well as non-parametric continuous models\cite{Nefian2006-nr}. Models that assume parallel lanes have been used\cite{Labayrade2006-ta}, and more recently models integrating topological elements such as lane splitting and merging were proposed\cite{Huang2009-ge}.

\begin{figure*}
	\centering\includegraphics[width=0.8\textwidth]{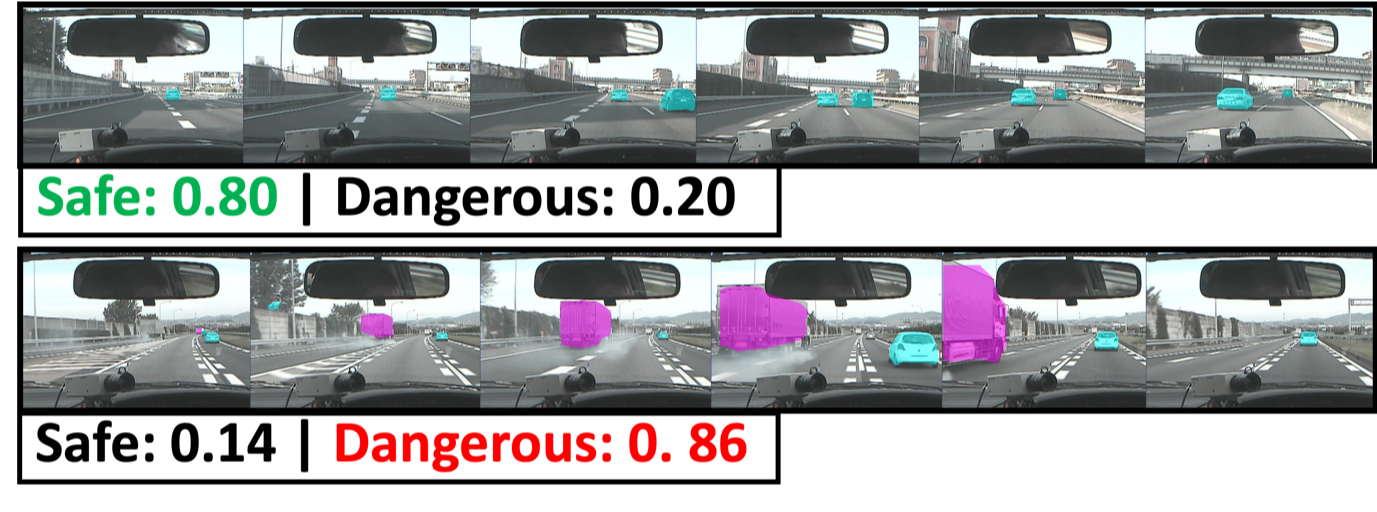}
	\caption{Assessing the overall risk level of driving scenes. We employed an open-source\textsuperscript{\ref{footnote1}} deep spatiotemporal video-based risk detection framework \cite{yurtsever2019risky} to assess the image sequences shown in this figure. }
	\label{fig_risk_assesment}
\end{figure*}


Temporal integration completes the road and lane segmentation pipeline. Here, vehicle dynamics are used in combination with a road tracking system to achieve smooth results. Dynamic information can also be used alongside Kalman filtering\cite{Labayrade2006-ta} or particle filtering\cite{Danescu2009-qr} to achieve smoother results.


Road and lane estimation is a well-researched field and many methods have already been integrated successfully for lane keeping assistance systems. However, most methods remain riddled with assumptions and limitations, and truly general systems which can handle complex road topologies have yet to be developed. Through standardized road maps which encode topology and emerging machine learning-based road and lane classification methods, robust systems for driving automation are slowly taking shape.



\section{Assessment}\label{sec_assessment}

A robust ADS should constantly evaluate the overall risk level of the situation and predict the intentions of surrounding human drivers and pedestrians. A lack of acute assessment mechanism can lead to accidents. This section discusses assessment under three subcategories: overall risk and uncertainty assessment, human driving behavior assessment, and driving style recognition.

\subsection{Risk and uncertainty assessment }

Overall assessment can be summarized as quantifying the uncertainties and the risk level of the driving scene. It is a promising methodology that can increase the safety of ADS pipelines \cite{mcallister2017concrete}. 

Using Bayesian methods to quantify and measure uncertainties of deep neural networks was proposed in \cite{gal2016uncertainty}. A Bayesian deep learning architecture was designed for propagating uncertainty throughout an ADS pipeline, and the advantage of it over conventional approaches was shown in a hypothetical scenario  \cite{mcallister2017concrete}. In summary, each module conveys and accepts probability distributions instead of exact outcomes throughout the pipeline, which increases the overall robustness of the system.     

An alternative approach is to assess the overall risk level of the driving scene separately, i.e outside the pipeline. Sensory inputs were fed into a risk inference framework in \cite{yurtsever2018integrating, yamazaki2016integrating} to detect unsafe lane change events using Hidden Markov Models (HMMs) and language models. Recently, a deep spatiotemporal network that infers the overall risk level of a driving scene was introduced in \cite{yurtsever2019risky}. Implementation of this method is available open-source\footnote{\label{footnote1}\href{https://github.com/Ekim-Yurtsever/DeepTL-Lane-Change-Classification}{https://github.com/Ekim-Yurtsever/DeepTL-Lane-Change-Classification}}. We employed this method to assess the risk level of a lane change as shown in Figure \ref{fig_risk_assesment}.

\subsection{Surrounding driving behavior assessment}


Understanding surrounding human driver intention  is most relevant to medium to long term prediction and decision making. In order to increase the prediction horizon of surrounding object behavior, human traits should be considered and incorporated into the prediction and evaluation steps. Understanding surrounding driver intention from the perspective of an ADS is not a common practice in the field, as such, state-of-the-art is not established yet. 



In \cite{geng2017scenario}, a target vehicle's future behavior was predicted with a hidden Markov model (HMM) and the prediction time horizon was extended 56\% by learning human driving traits. The proposed system tagged observations with predefined maneuvers. Then, the features of each type were learned in a data-centric manner with HMMs. Another learning based approach was proposed in \cite{bahram2016combined}, where a Bayesian network classifier was used to predict maneuvers of individual drivers on highways. A framework for long term driver behavior prediction using a combination of a hybrid state system and HMM was introduced in \cite{gadepally2017framework}. Surrounding vehicle information was integrated with ego-behavior through a symbolization framework in \cite{yamazaki2016integrating, yurtsever2018integrating}. Detecting dangerous cut in maneuvers was achieved with an HMM framework that was trained on safe and dangerous data in \cite{liu2015classification}. Lane change events were predicted 1.3 seconds in advance with support vector machines (SVM) and Bayesian filters \cite{kumar2013learning}.
 
The main challenges are the short observation window for understanding the intention of humans and real-time high-frequency computation requirements. ADSs can typically only observe a surrounding vehicle only for seconds. Complicated driving behavior models that require longer observation periods cannot be utilized under these circumstances.

\begin{figure*}
	\centering\includegraphics[width=1.8\columnwidth]{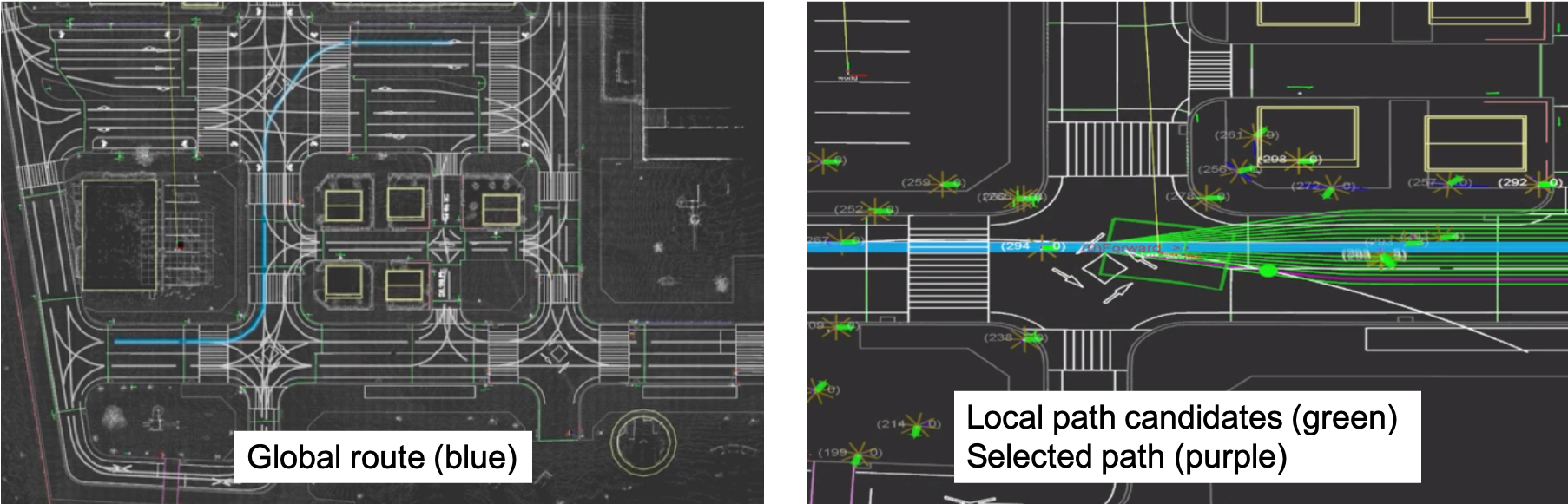}
	\caption{Global plan and the local paths. The annotated vector map shown in \Cref{fig_vector_map} was utilized by the planner. We employed OpenPlanner \cite{darweesh2017open}, which is a graph-based planner, to illustrate a typical planning approach.}
	\label{fig_planning}
\end{figure*}

\subsection{Driving style recognition} \label{sec_driving_behavior}



In 2016, Google's self-driving car collided with an oncoming bus \cite{davies2016google} during a lane change where it assumed that the bus driver was going to yield. However, the bus driver accelerated instead. This accident may have been prevented if the ADS understood the bus driver's individual, unique driving style and predicted his behavior.

Driving style is a broad term without an established common definition. Furthermore, recognizing the surrounding human driving styles is a severely understudied topic. However, thorough reviews of driving style categorization of \textit{human-driven} ego vehicles can be found in \cite{sagberg2015review} and in \cite{martinez2018driving}. Readers are referred to these papers for a complete review. The remainder of this subsection gives a brief overview of \textit{human-driven} ego vehicle-based driving style recognition.

Typically, driving style is defined with respect to either aggressiveness \cite{johnson2011driving, fazeen2012safe, karginova2012data, doshi2010examining, vaitkus2014driving} or fuel consumption \cite{syed2010design, corti2013quantitative, ericsson2001independent, manzoni2010driving, neubauer2013accounting}. For example, \cite{murphey2009driver} introduced a rule-based model that classified driving styles with respect to jerk. This model decides whether a maneuver is aggressive or calm by a set of rules and jerk thresholds.  Drivers were categorized with respect to their average speed in \cite{yurtsever2015traffic}. In conventional methods, total number and meaning of driving style classes are predefined beforehand. The vast majority of driving style recognition literature uses two \cite{johnson2011driving, syed2010design, fazeen2012safe, yamazaki2016integrating, yurtsever2018integrating} or three \cite{dorr2014online, xu2015establishing, rajan2012investigation} classes. Representing driving style in a continuous domain is uncommon, but there are some studies. In \cite{augustynowicz2009preliminary}, driving style was depicted as a continuous value between -1 and +1, which stands for mild and active respectively. Details of classification methods are given in Table \ref{table_survey_driving_style}.


\begin{table}
	\footnotesize
	\renewcommand{\arraystretch}{1.3}
	\caption{Driving style categorization of human-driven ego vehicles}
	\label{table_survey_driving_style}
	\centering
	\begin{tabularx}{\columnwidth}{@{}cccY@{}}
		\specialrule{.1em}{.05em}{.05em} 
		\B \T	
		\textbf{Related work} & \textbf{\# Classes} & \textbf{Methodology} & \textbf{Class details}\T\\
		\hline
		\cite{constantinescu2010driving} & 5 & PCA & non-aggressive to very aggressive \\
		\cite{zhang2010pattern} & 3 & NN, SVM, DT & expert/typical/low-skill \\
		\cite{dorr2014online}& 3 & FL & sporty/normal/comfortable \\
		\cite{xu2015establishing}& 3 & PCMLP & aggressive/moderate/calm \\    
		\cite{yurtsever2019traffic} & 3 & \thead{SAE \& K-means} & \thead{unidentified clusters } \\ 
		\cite{yurtsever2018integrating} & 2 & \thead{non-param. \\ Bayesian} & risky/safe \\
		\cite{johnson2011driving}& 2 & DTW & aggressive/non-aggressive  \\    
		\cite{fazeen2012safe}& 2 & RB & sudden/safe \\    
		\cite{augustynowicz2009preliminary} & \thead{Continuous \\ $[-1,1]$} & NN & mild to active \\
		\specialrule{.1em}{.05em}{.05em} 
	\end{tabularx}
\end{table}

More recently, machine learning based approaches have been utilized for driving style recognition. Principal component analysis was used and five distinct driving classes were detected in an unsupervised manner in \cite{constantinescu2010driving} and a GMM based driver model was used to identify individual drivers with success in \cite{miyajima2007driver}. Car-following and pedal operation behavior was investigated separately in the latter study. Another GMM based driving style recognition model was proposed for electric vehicle range prediction in \cite{bolovinou2014online}. In \cite{johnson2011driving}, aggressive event detection with dynamic time warping was presented where the authors reported a high success score. Bayesian approaches were utilized in \cite{mudgal2014driving} for modeling driving style on roundabouts and in \cite{mccall2007driver} to asses critical braking situations. Bag-of-words and K-means clustering was used to represent individual driving features in \cite{yurtsever2015driving}. A stacked autoencoder was used to extract unique driving signatures from different drivers, and then macro driving style centroids were found with clustering \cite{yurtsever2019traffic}. Another autoencoder network was used to extract road-type specific driving features \cite{sama2018driving}. Similarly, driving behavior was encoded in a 3-channel RGB space with a deep sparse autoencoder to visualize individual driving styles \cite{liu2017visualization}.

A successful integration of driving style recognition into a real world ADS pipeline is not reported yet. However, these studies are promising and point to a possible new direction in ADS development.

\section{Planning and decision making}\label{sec_planning}

Planning can be divided into two sub-tasks: global route planning and local path planning. Figure \ref{fig_planning} illustrates a typical planning approach in detail. 

The remainder of this section gives a brief overview of the subject. For more information studies such as \cite{paden2016survey, bast2016route, badue2019self} can be referred.

\subsection{Global planning}

The global planner is responsible for finding the route on the road network from origin to the final destination. The user usually defines the final destination. Global navigation is a well-studied subject, and high performance has become an industry standard for more than a decade. Almost all modern production cars are equipped with navigation systems that utilize GPS and offline maps to plan a global route.

Route planning is formulated as finding the point-to-point shortest path in a directed graph, and conventional methods are examined under four categories in \cite{bast2016route}. These are; goal-directed, separator-based, hierarchical and bounded-hop techniques. A* search \cite{hart1968formal} is a standard goal-directed path planning algorithm and used extensively in various fields for almost 50 years. 

The main idea of separator-based techniques is to remove a subset of vertices \cite{van1978improved} or arcs from the graph and compute an overlay graph over it. Using the overlayed graph to calculate the shortest path results in faster queries. 

Hierarchical techniques take advantage of the road hierarchy. For example, the road hierarchy in the US can be listed from top to bottom as freeways, arterials, collectors and local roads respectively. For a route query, the importance of hierarchy increases as the distance between origin and destination gets longer. The shortest path may not be the fastest nor the most desirable route anymore. Getting away from the destination thus making the route a bit longer to take the closest highway ramp may result in faster travel time in comparison to following the shortest path of local roads. Contraction Hierarchies (CH) method  was proposed in \cite{geisberger2012exact} for exploiting road hierarchy. 

Precomputing distances between selected vertexes and utilizing them on the query time is the basis of bounded-hop techniques. Precomputed shortcuts can be utilized partly or exclusively for navigation. However, the naive approach of precomputing all possible routes from every pair of vertices is impractical in most cases with large networks. One possible solution to this is to use hub labeling (HL) \cite{cohen2003reachability}. This approach requires preprocessing also. A label associated with a vertex consists of nearby \textit{hub} vertices and the distance to them. These labels satisfy the condition that at least one shared hub vertex must exist between the labels of any given two vertices. HL is the fastest query time algorithm for route planning \cite{bast2016route}, in the expense of high storage usage.

A combination of the above algorithms are popular in state-of-the-art systems. For example, \cite{bauer2010combining} combined a separator with a bounded-hop method and created the Transit Node Routing with Arc Flags (TNR+AF) algorithm. Modern route planners can make a query in milliseconds. 

\subsection{Local planning}

The objective of the local planner is to execute a global plan without \textit{failing}. In other words, in order to complete its trip, the ADS must find trajectories to avoid obstacles  and satisfy optimization criteria in the configuration space (C-space), given a starting and destination point. A detailed local planning review is presented in  \cite{gonzalez2016review} where the taxonomy of motion planning was divided into four groups; graph-based planners, sampling-based planners, interpolating curve planners and numerical optimization approaches. After a summary of these conventional planners, the emerging deep learning-based planners are introduced at the end of this section. Table \ref{table_path_planning} gives a brief summary of local planning methods.  

\begin{table}
	\footnotesize
	\centering
	\renewcommand{\arraystretch}{1.3}
	\caption{Local planning techniques}
	\label{table_path_planning}
	{\begin{tabular}{ccc}
			
			\specialrule{.1em}{.05em}{.05em} 
			\B \T
			Approach & Methods & \thead{Pros and cons}  \\
			\hline
			\thead{Graph \\ search} & \thead{Dijkstra\cite{delling2013phast}, A*\cite{hart1968formal}, \\ State lattice\cite{pivtoraiko2005efficient}} & Slow and jerky \\
			\thead{Sampling \\ based} & \thead{RPP\cite{barraquand1991robot}, RRT\cite{lavalle2001randomized}, \\ RRT*\cite{karaman2011sampling}, PRM\cite{kavraki1994probabilistic}}  & \thead{Fast solution but jerky} \\
			\thead{Curve \\ interpolation} & \thead{clothoids\cite{fuji2014trajectory}, \\ polynomials\cite{petrov2014modeling}, \\ Bezier\cite{rastelli2014dynamic}, splines\cite{bergenhem2012overview}} & Smooth but slow \\
			\thead{Numerical \\ optimization} & \thead{num. non-linear opt. \cite{dolgov2010path}, \\ Newton's method \cite{ren2006modified}}  & \thead{increases computational \\ cost but improves quality}  \\
			\thead{Deep \\ learning} & \thead{FCN \cite{caltagirone2017lidar}, \\segmentation network \cite{barnes2017find}} & \thead{high imitation performance, \\but no hard coded \\ safety measures} \\

			\specialrule{.1em}{.05em}{.05em} 
	\end{tabular}}{}
\end{table}

Graph-based local planners use the same techniques as graph-based global planners such as Dijkstra \cite{delling2013phast} and A* \cite{hart1968formal}, which output discrete paths rather than continuous ones. This can lead to jerky trajectories \cite{gonzalez2016review}. A more advanced graph-based planner is the state lattice algorithm. As all graph-based methods, state lattice discretizes the decision space. High dimensional lattice nodes, which typically encode 2D position, heading and curvature \cite{pivtoraiko2005efficient}, are used to create a grid first. Then, the connections between the nodes are precomputed with an inverse path generator to build the state lattice. During the planning phase, a cost function, which usually considers proximity to obstacles and deviation from the goal, is utilized for finding the best path with the precomputed path primitives. State lattices can handle high dimensions and is good for local planning in dynamical environments, however, the computational load is high and the discretization resolution limits the planners' capacity \cite{gonzalez2016review}.          


A detailed overview of Sampling Based Planning (SBP) methods can be found in \cite{elbanhawi2014sampling}. In summary, SBP tries to build the connectivity of the C-space by randomly sampling paths in it. Randomized Potential Planner (RPP) \cite{barraquand1991robot} is one of the earliest SBP approaches, where random walks are generated for escaping local minimums. Probabilistic roadmap method (PRM) \cite{kavraki1994probabilistic} and rapidly-exploring random tree (RRT) \cite{lavalle2001randomized} are the most commonly used SBP algorithms. PRM first samples the C-space during its learning phase and then makes a query with the predefined origin and destination points on the roadmap. RRT, on the other hand, is a single query planner. The path between start and goal configuration is incrementally built with random tree-like branches. RRT is faster than PRM and both are probabilistically complete \cite{lavalle2001randomized}, which means a path that satisfies the given conditions will be guaranteed to be found with \textit{enough} runtime. RRT* \cite{karaman2011sampling}, an extension of the RRT, provides more optimal paths instead of completely random ones while sacrificing computational efficiency. The main disadvantage of SBP in general is, again, the jerky trajectories \cite{gonzalez2016review}. 

Interpolating curve planners fit a curve to a known set of points \cite{gonzalez2016review}, e.g. way-points generated from the global plan or a discrete set of future points from another local planner.  The main obstacle avoidance strategy is to interpolate new collision-free paths that first deviate from, and then re-enter back to the initial planned trajectory. The new path is generated by fitting a curve to a new set of points: an exit point from the currently traversed trajectory, newly sampled collision free points, and a re-entry point on the initial trajectory.
The resultant trajectory is smooth, however, the computational load is usually higher compared to other methods. There are various curve families that are used commonly such as clothoids \cite{fuji2014trajectory}, polynomials \cite{petrov2014modeling}, Bezier curves \cite{rastelli2014dynamic} and splines \cite{bergenhem2012overview}.

Optimization based motion planners improve the quality of already existing paths with optimization functions. A* trajectories were optimized with numeric non-linear functions in \cite{dolgov2010path}. Potential Field Method (PFM) was improved by solving the inherent oscillation problem using Newton's method with obtaining $C_{1}$ continuity in \cite{ren2006modified}.   

Recently, Deep Learning (DL) and reinforcement learning based local planners started to emerge as an alternative. Fully convolutional 3D neural networks can generate future paths from sensory input such as lidar point clouds \cite{caltagirone2017lidar}. An interesting take on the subject is to segment image data with path proposals using a deep segmentation network \cite{barnes2017find}. Planning a safe path in occluded intersections was achieved in a simulation environment using deep reinforcement learning in \cite{isele2018navigating}.   The main difference between end-to-end driving and deep learning based local planners is the output: the former outputs direct vehicle control signals such as steering and pedal operation, whereas the latter generates a trajectory. This enables DL planners to be integrated into conventional pipelines \cite{schwarting2018planning}. 

Deep learning based planners are promising, but they are not widely used in real-world systems yet. Lack of hard-coded safety measures, generalization issues, need for labeled data are some of the issues that need to be addressed.

\section{Human machine interaction}\label{sec_HMI}


Vehicles communicate with their drivers/passengers through their HMI module. The nature of this communication greatly depends on the objective, which can be divided into two: primary driving tasks  and secondary tasks. The interaction intensity of these tasks depend on the automation level. Where a manually operated, level zero conventional car requires constant user input for operation, a level five ADS may need user input only at the beginning of the trip. Furthermore, the purpose of interaction may affect intensity. A shift from executing primary driving tasks to monitoring the automation process raises new HMI design requirements.      


There are several investigations such as \cite{pickering2007review, carsten2019can} about automotive HMI technologies, mostly from the distraction point of view. Manual user interfaces for secondary tasks are more desired than their visual counterparts \cite{pickering2007review}. The main reason is vision is absolutely necessary and has no alternative for primary driving tasks. Visual interface interactions require glances with durations between 0.6 and 1.6 seconds with a mean of 1.2 seconds\cite{pickering2007review}. As such, secondary task interfaces that require vision is distracting and detrimental for driving.  



Auditory User Interfaces (AUI) are good alternatives to visually taxing HMI designs. AUIs are omni-directional: even if the user is not attending, the auditory cues are hard to miss \cite{bazilinskyy2015auditory}. The main challenge of audio interaction is automatic speech recognition (ASR). ASR is a very mature field. However, in vehicle domain there are additional challenges; low performance caused by uncontrollable cabin conditions such as wind and road noise \cite{peden2004world}. Beyond simple voice commands, conversational natural language interaction with an ADS is still an unrealized concept with many unsolved challenges \cite{large2017steering}. 

\begin{table*}[!t]
	\caption{Driving datasets}
	\label{table_datasets}
	{\begin{tabularx}{\textwidth}{@{}cccccccYYYY@{}}
			\specialrule{.1em}{.05em}{.05em} 
			Dataset & Image & LIDAR & \thead{2D \\ annotation*} & \thead{3D \\ annotation* } & ego signals & Naturalistic & POV & \thead{Multi \\ trip} & \thead{all \\ weathers}  & \thead{day \&\\ night}\\ 
			\hline 
			Cityscapes  \cite{cordts2016cityscapes} & \checkmark & - & \checkmark & -  & - & - &Vehicle & - & - & -\\
			Berkley DeepDrive \cite{yu2018bdd100k} & \checkmark  & - & \checkmark & - & - & - &Vehicle & - & \checkmark & \checkmark\\
			Mapillary \cite{neuhold2017mapillary} & \checkmark & - & \checkmark & -& - & - &Vehicle & - & \checkmark & \checkmark\\    
			Oxford RobotCar \cite{maddern20171} & \checkmark & \checkmark & - & - & - & - & Vehicle & \checkmark  & \checkmark & \checkmark\\
			KITTI \cite{geiger2012we} & \checkmark & \checkmark & \checkmark & \checkmark & -  & - &Vehicle & - & - & -\\
			H3D \cite{patil2019h3d} & \checkmark & \checkmark & - & \checkmark & - & - & Vehicle & -  & - & -\\
			ApolloScape \cite{huang2018apolloscape} & \checkmark &\checkmark & \checkmark & \checkmark & -  & - &Vehicle & - & - & -\\
			nuScenes \cite{nuScene2018} & \checkmark &\checkmark & \checkmark & \checkmark & -  & - &Vehicle & - & \checkmark & \checkmark\\
			Udacity \cite{udacity}  & \checkmark &\checkmark & \checkmark & \checkmark & -  & - &Vehicle & - & - & -\\
			DDD17 \cite{binas2017ddd17} & \checkmark & - & \checkmark & - & \checkmark  & - & Vehicle & - & \checkmark & \checkmark\\
			Comma2k19 \cite{comma2k19} & \checkmark & - & - & - & \checkmark & - & Vehicle & - & - & \checkmark\\
			LiVi-Set \cite{chen2018lidar} & \checkmark & \checkmark & - & - &  \checkmark & - & Vehicle & - & - & -\\
			
			NU-drive \cite{takeda2011international} & \checkmark & - & - & - & \checkmark & Semi & Vehicle & \checkmark  & - & -\\
			SHRP2 \cite{blatt2015naturalistic} & \checkmark & - & - & - & \checkmark & \checkmark & Vehicle & - & - & -\\
			100-Car \cite{klauer2010analysis} & \checkmark & - & - & - & \checkmark & \checkmark & Vehicle & - & \checkmark & \checkmark\\
			euroFOT \cite{benmimoun2013eurofot} & \checkmark & - & - & - & \checkmark & \checkmark & Vehicle& -  & - & -\\
			TorontoCity \cite{wang2017torontocity} & \checkmark & \checkmark & \checkmark & \checkmark & - & - & \thead{Aerial, \\panorama, \\vehicle}& - & - & -\\
			
			KAIST multi-spectral \cite{choi2018kaist} & \checkmark & \checkmark & \checkmark & - & - & - & Vehicle & - & - & \checkmark\\
			\specialrule{.1em}{.05em}{.05em} 
			\multicolumn{11}{l}{*2D and 3D annotation can vary from bounding boxes to segmentation masks. Readers are referred to sources for details of the datasets.}
		\end{tabularx}}{}
		\vspace{-0.5cm}

	\end{table*}

The biggest HMI challenge is at automation level three and four. The user and the ADS need to have a mutual understanding, otherwise, they will not be able to grasp each other's intentions  \cite{carsten2019can}. The transition from manual to automated driving and vice versa is prone to fail in the state-of-the-art. Recent research showed that drivers exhibit low cognitive load when monitoring automated driving compared to doing a secondary task \cite{sibi2016monitoring}. Even though some experimental systems can recognize driver-activity with a driver facing camera based on head and eye-tracking \cite{braunagel2015driver}, and prepare the driver for handover with visual and auditory cues \cite{walch2015autonomous} in simulation environments, a real world system with an efficient handover interaction module does not exist at the moment. This is an open problem \cite{hansen2017driver} and future research should focus on delivering better methods to inform/prepare the driver for easing the transition \cite{merat2014transition}.


%
%
%

%
%

\section{Datasets and available tools}\label{sec_datasets}

\subsection{Datasets and Benchmarks}
									
Datasets are crucial for researchers and developers because most of the algorithms and tools have to be tested and trained before going on road. 

Typically, sensory inputs are fed into a stack of algorithms with various objectives. A common practice is to test and validate these functions separately on annotated datasets. For example, the output of cameras, 2D vision, can be fed into an object detection algorithm to detect surrounding vehicles and pedestrians. Then, this information can be used in another algorithm for planning purposes. Even though these two algorithms are connected in the stack of this example, the object detection part can be worked on and validated separately during the development process. Since computer vision is a well-studied field, there are annotated datasets for object detection and tracking specifically. The existence of these datasets increases the development process and enables interdisciplinary research teams to work with each other much more efficiently. For end-to-end driving, the dataset has to include additional ego-vehicle signals, chiefly steering and longitudinal control signals. 

As learning approaches emerged, so did training datasets to support them. The PASCAL VOC dataset\cite{Everingham2006-fs}, which grew from 2005 to 2012, was one of the first dataset featuring a large amount of data with relevant classes for ADS. However, the images often featured single objects, in scenes and scales that are not representative of what is encountered in driving scenarios. In 2012, the KITTI Vision Benchmark \cite{geiger2012we} remedied this situation by providing a relatively large amount of labeled driving scenes. It remains as one of the most widely used datasets for applications related to driving automation. Yet in terms of quantity of data and number of labeled classes, it is far inferior to generic image databases such as ImageNet\cite{Deng2009-yh} and COCO \cite{Lin2014-rh}. While no doubt useful for training, these generic image databases lack the adequate context to test the capabilities of ADS. UC Berkeley DeepDrive \cite{yu2018bdd100k} is a recent dataset with annotated image data. The Oxford RobotCar \cite{maddern20171} was used to collect over 1000 km of driving data with six cameras, lidar, GPS and INS in the UK but is not annotated. ApolloScape is a very recent dataset that is not fully public yet   \cite{huang2018apolloscape}. Cityscapes \cite{cordts2016cityscapes} is commonly used for computer vision algorithms as a benchmark set. Mapillary Vistas is a big image dataset with annotations \cite{neuhold2017mapillary}. TorontoCity benchmark \cite{wang2017torontocity} is a very detailed dataset; however it is not public yet. The nuScenes dataset is the most recent urban driving dataset with lidar and image sensors \cite{nuScene2018}. Comma.ai has released a part of their dataset \cite{santana2016learning} which includes 7.25 hours of driving. In DDD17 \cite{binas2017ddd17} around 12 hours of driving data is recorded. The LiVi-Set \cite{chen2018lidar} is a new dataset that has lidar, image and driving behavior. CommonRoad \cite{althoff2017commonroad} is a new benchmark for motion-planning.


Naturalistic driving data is another type of dataset that concentrates on the individual element of the driving: the driver. SHRP2 \cite{blatt2015naturalistic} includes over 3000 volunteer participants' driving data over a 3-year collection period. Other naturalistic driving datasets are the 100-Car study \cite{klauer2010analysis}, euroFOT \cite{benmimoun2013eurofot} and NUDrive \cite{takeda2011international}.   Table \ref{table_datasets} shows the comparison of these datasets.

\subsection{Open-source frameworks and simulators}

Open source frameworks are very useful for both researchers and the industry. These frameworks can ``democratize" ADS development. Autoware \cite{kato2015open}, Apollo \cite{fan2018baidu}, Nvidia DriveWorks \cite{nvidia2018} and openpilot \cite{commaai2018} are amongst the most used software-stacks capable of running an ADS platform in real world. We utilized Autoware \cite{kato2015open} to realize core automated driving functions in this study.


Simulations also have an important place for ADS development. Since the instrumentation of an experimental vehicle still has a high cost and conducting experiments on public road networks are highly regulated, a simulation environment is beneficial for developing certain algorithms/modules before road tests. Furthermore, highly dangerous scenarios such as a collision with pedestrian can be tested in simulations with ease. CARLA \cite{dosovitskiy2017carla} is an urban driving simulator developed for this purpose. TORCS \cite{wymann2000torcs} was developed for race track simulation. Some researchers even used computer games such as Grand Theft Auto V \cite{richter2017playing}. Gazebo \cite{koenig2004design} is a common simulation environment for robotics. For traffic simulations, SUMO \cite{krajzewicz2002sumo} is a widely used open-source platform. \cite{stellet2015testing} proposed different concepts of integrating real-world measurements into the simulation environment.

\section{Conclusions}\label{chap_conclusion}
In this survey on automated driving systems, we outlined some of the key innovations as well as existing systems. While the promise of automated driving is enticing and already marketed to consumers, this survey has shown there remains clear gaps in the research. Several architecture models have been proposed, from fully modular to completely end-to-end, each with their own shortcomings. The optimal sensing modality for localization, mapping and perception is still disagreed upon, algorithms still lack accuracy and efficiency, and the need for a proper online assessment has become apparent. Less than ideal road conditions are still an open problem, as well as dealing with intemperate weather. Vehicle-to-vehicle communication is still in its infancy, while centralized, cloud-based information management has yet to be implemented due to the complex infrastructure required. Human-machine interaction is an under-researched field with many open problems.


The development of automated driving systems relies on the advancements of both scientific disciplines and new technologies. As such, we discussed the recent research developments which are likely to have a significant impact on automated driving technology, either by overcoming the weakness of previous methods or by proposing an alternative. This survey has shown that through inter-disciplinary academic collaboration and support from industries and the general public, the remaining challenges can be addressed. With directed efforts towards ensuring robustness at all levels of automated driving systems, safe and efficient roads are just beyond the horizon.  \rule{0.8ex}{1.2ex}

%

%
%
%
%

\bibliographystyle{IEEEtran}
\bibliography{perception,references_auto_survey}

\vspace{-1cm}

\begin{IEEEbiography}[{\includegraphics[width=1in,height=1.25in,clip,keepaspectratio]{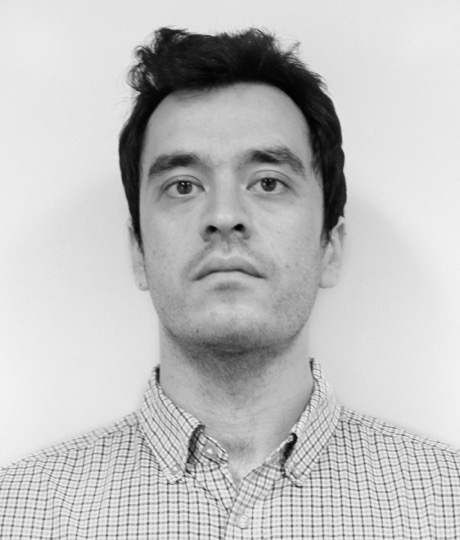}}]{Ekim Yurtsever} (Member, IEEE)
	received his B.S. and M.S. degrees from Istanbul Technical University in 2012 and 2014 respectively. He received his Ph.D. in Information Science in 2019 from Nagoya University, Japan and is working as a postdoctoral researcher at the Department of Electrical and Computer Engineering, Ohio State University since 2019. 
	
	His research interests include artificial intelligence, machine learning, and computer vision. Currently, he is working on machine learning and computer vision tasks in the intelligent vehicle domain.

\end{IEEEbiography}

\begin{IEEEbiography}[{\includegraphics[width=1in,height=1.25in,clip,keepaspectratio]{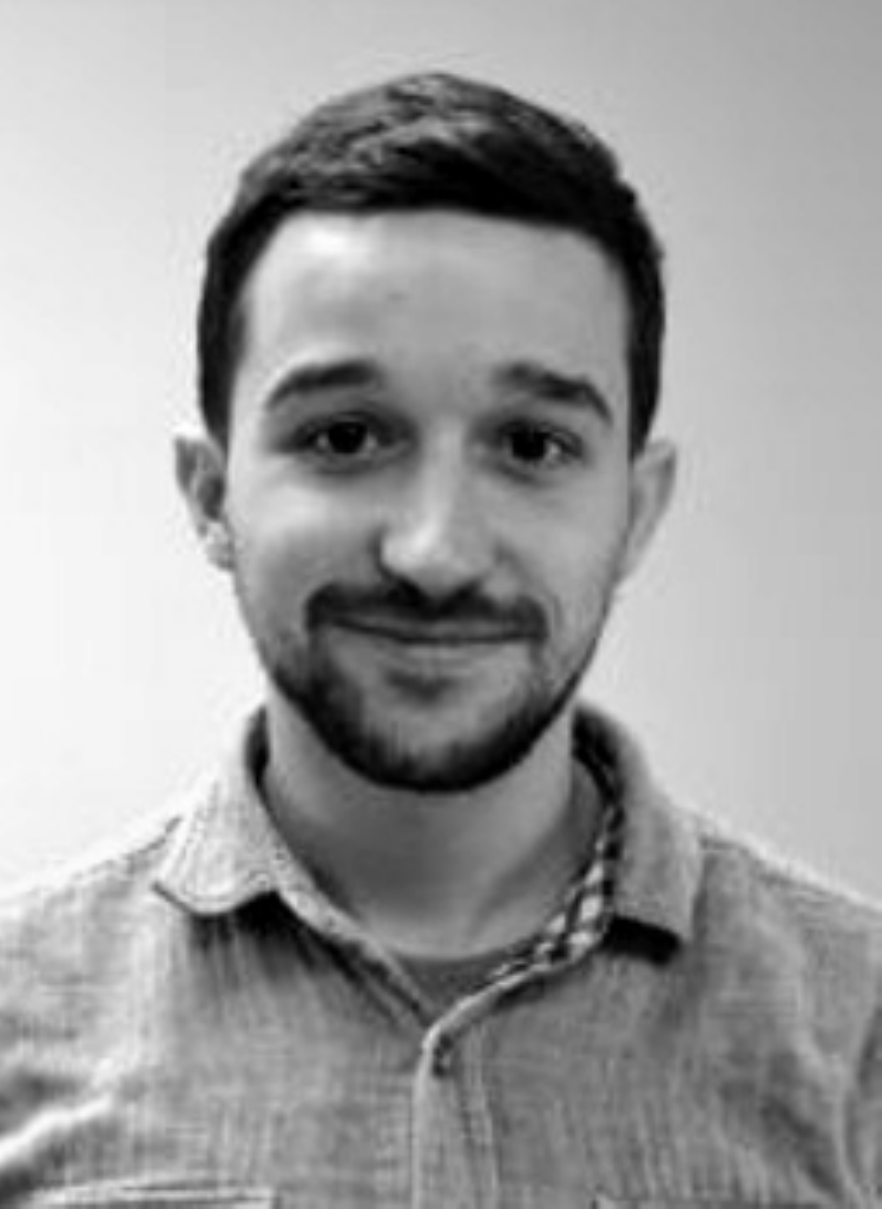}}]{Jacob Lambert} (Student Member, IEEE)
	received his B.S. in Honours Physics in 2014 at McGill University in Montreal, Canada. He received his M.A.Sc. in 2017 at the University of Toronto, Canada, and is currently a PhD candidate in Nagoya University, Japan. 
	
	His current research focuses on 3D perception through lidar sensors for autonomous robotics.

\end{IEEEbiography}

\begin{IEEEbiography}[{\includegraphics[width=1in,height=1.25in,clip,keepaspectratio]{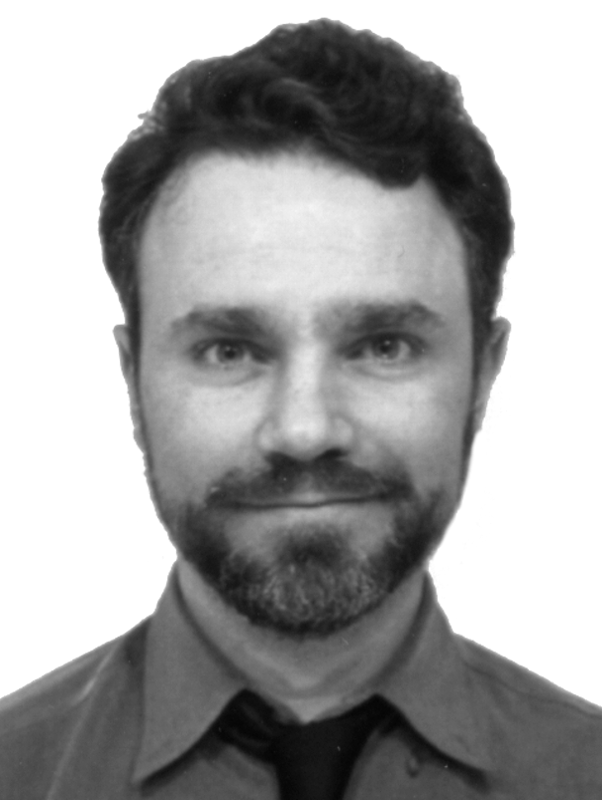}}]{Alexander Carballo} (Member, IEEE)
	received his Dr.Eng. degree from the Intelligent Robot Laboratory, University of Tsukuba, Japan. From 1996 to 2006, he worked as lecturer at School of Computer Engineering, Costa Rica Institute of Technology. From 2011 to 2017, worked in Research and Development at Hokuyo Automatic Co., Ltd. Since 2017, he is a Designated Assistant Professor at Institutes of Innovation for Future Society, Nagoya University, Japan. 
	
	His main research interests are lidar sensors, robotic perception and autonomous driving.

\end{IEEEbiography}

\begin{IEEEbiography}[{\includegraphics[width=1in,height=1.25in,clip,keepaspectratio]{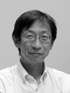}}]{Kazuya Takeda} (Senior Member, IEEE)
	received his B.E.E., M.E.E., and Ph.D. from Nagoya University, Japan. Since 1985 he had been working at Advanced Telecommunication Research Laboratories and at KDD R\&D Laboratories, Japan. In 1995, he started a research group for signal processing applications at Nagoya University. 

	He is currently a Professor at the Institutes of Innovation for Future Society, Nagoya University and with Tier IV inc. He is also serving as a member of the Board of Governors of the IEEE ITS society.
	
	His main focus is investigating driving behavior using data centric approaches, utilizing signal corpora of real driving behavior.

\end{IEEEbiography}

\EOD

\end{document}